\documentclass[letterpaper]{article} % DO NOT CHANGE THIS
\usepackage{aaai2026}  % DO NOT CHANGE THIS
\usepackage{times}  % DO NOT CHANGE THIS
\usepackage{helvet}  % DO NOT CHANGE THIS
\usepackage{courier}  % DO NOT CHANGE THIS
\usepackage[hyphens]{url}  % DO NOT CHANGE THIS
\usepackage{graphicx} % DO NOT CHANGE THIS
\usepackage{natbib}  % DO NOT CHANGE THIS AND DO NOT ADD ANY OPTIONS TO IT
\usepackage{caption} % DO NOT CHANGE THIS AND DO NOT ADD ANY OPTIONS TO IT
\usepackage{algorithm}
\usepackage{algorithmic}

\usepackage{booktabs}       % professional-quality tables
\usepackage{amsfonts}       % blackboard math symbols
\usepackage{nicefrac}       % compact symbols for 1/2, etc.
\usepackage{subfigure}
\usepackage{bm}
\usepackage{bbold}
\usepackage{amsmath}
\usepackage{amssymb}
\usepackage{mathtools}
\usepackage{amsthm}

\theoremstyle{plain}
\newtheorem{theorem}{Theorem} %[section]

\theoremstyle{definition}
\newtheorem{definition}[theorem]{Definition}

\theoremstyle{remark}

\newcommand{\norm}[1]{\left\lVert#1\right\rVert}

\newcommand{\prox}{\operatorname{prox}}
\newcommand{\trace}{\operatorname{trace}}
\DeclareMathOperator*{\argmin}{argmin}
\DeclareMathOperator*{\argmax}{argmax}
\usepackage{xspace}
\usepackage{cancel}

\usepackage{pythonhighlight}
\usepackage{multirow}
\newcommand{\STAB}[1]{\begin{tabular}{@{}c@{}}#1\end{tabular}}

\usepackage{xpatch}
\xpatchcmd{\boxed}{%
\fbox}{%
\fcolorbox{red}{white}}{}{}
\makeatletter
\DeclareRobustCommand\onedot{\futurelet\@let@token\@onedot}
\def\@onedot{\ifx\@let@token.\else.\null\fi\xspace}

\def\eg{\emph{e.g}\onedot} 
\def\ie{\emph{i.e}\onedot}

\let\titleold\title
\renewcommand{\title}[1]{\titleold{#1}\newcommand{\thetitle}{#1}}
\def\maketitlesupplementary
   {
   \newpage
       \twocolumn[
        \centering
        \Large
        \textbf{\thetitle}\\
        \vspace{0.5em}Appendix \\
        \vspace{1.0em}
       ] %< twocolumn
   }

\makeatother

\usepackage{xcolor}

\usepackage{newfloat}
\usepackage{listings}
\DeclareCaptionStyle{ruled}{labelfont=normalfont,labelsep=colon,strut=off} % DO NOT CHANGE THIS
\floatstyle{ruled}
\newfloat{listing}{tb}{lst}{}
\floatname{listing}{Listing}
\title{Image Restoration via Primal Dual Hybrid Gradient and Flow Generative Model}
\author{
    Ji Li\textsuperscript{\rm 1},
    Chao Wang\textsuperscript{\rm 2}\thanks{Corresponding author.}
}
\affiliations{
    \textsuperscript{\rm 1}Academy for Multidisciplinary Studies, Capital Normal University, Beijing, China\\
    \textsuperscript{\rm 2}College of Science, China Agricultural University, Beijing, China\\
    matliji@163.com, wywwwnx@163.com
}

\begin{document}

\maketitle

\begin{abstract}
Regularized optimization has been a classical approach to solving imaging inverse problems, where the regularization term enforces desirable properties of the unknown image. Recently, the integration of flow matching generative models into image restoration has garnered significant attention, owing to their powerful prior modeling capabilities. In this work, we incorporate such generative priors into a Plug-and-Play (PnP) framework based on proximal splitting, where the proximal operator associated with the regularizer is replaced by a time-dependent denoiser derived from the generative model. While existing PnP methods have achieved notable success in inverse problems with smooth squared $\ell_2$ data fidelity—typically associated with Gaussian noise—their applicability to more general data fidelity terms remains underexplored. To address this, we propose a general and efficient PnP algorithm inspired by the primal-dual hybrid gradient (PDHG) method. Our approach is computationally efficient, memory-friendly, and accommodates a wide range of fidelity terms. In particular, it supports both $\ell_1$ and $\ell_2$ norm-based losses, enabling robustness to non-Gaussian noise types such as Poisson and impulse noise. We validate our method on several image restoration tasks, including denoising, super-resolution, deblurring, and inpainting, and demonstrate that $\ell_1$ and $\ell_2$ fidelity terms outperform the conventional squared $\ell_2$ loss in the presence of non-Gaussian noise.
\end{abstract}

\setlength{\abovedisplayskip}{3pt}
\setlength{\belowdisplayskip}{3pt}

% Uncomment the following to link to your code, datasets, an extended version or similar.
% You must keep this block between (not within) the abstract and the main body of the paper.
% \begin{links}
%     \link{Code}{https://aaai.org/example/code}
%     \link{Datasets}{https://aaai.org/example/datasets}
%     \link{Extended version}{https://aaai.org/example/extended-version}
% \end{links}

\section{Introduction}

Many image restoration tasks in image processing and computer vision can be formulated as solving a linear inverse problem:
\begin{equation}
  \label{eq:prob}
  \bm{y} = \text{noisy}(\bm{A}\bm{x}),
\end{equation}
where $\bm{x}\in\mathbb{R}^n$ denotes the unknown clean image, $\bm{y}\in\mathbb{R}^m$ and linear matrix $\bm{A}\in\mathbb{R}^{m\times n}$ describes the degradation model. Image restoration is to recover $\bm{x}$ from the noisy observation $\bm{y}$.

In general, problem~\eqref{eq:prob} is ill-posed, and direct inversion is infeasible without incorporating prior knowledge about the solution. A widely adopted approach is to compute the maximum a posteriori (MAP) estimate of the posterior distribution $p(\bm{x}|\bm{y})$. The MAP estimate corresponds to the most probable solution and is given by:
\begin{equation}
  \label{eq:map}
  \hat{\bm{x}}=\argmax_{\bm{x}\in\mathbb{R}^n}\quad \{\log p(\bm{y}|\bm{x}) + \log p(\bm{x})\},
\end{equation}
where $p(\bm{y}|\bm{x})$ models the noise distribution, and $p(\bm{x})$ is the image prior. Since directly computing $\log p(\bm{x})$ is often intractable, MAP estimation is commonly reformulated as a regularized optimization problem:
\begin{equation}
  \label{eq:opt}
  \hat{\bm{x}}=\argmin_{\bm{x}\in\mathbb{R}^n}\quad F(\bm{A}\bm{x}) + G(\bm{x}),
\end{equation}
where $F(\cdot)$ encodes the data fidelity, measuring consistency between $\bm{A}\bm{x}$ and $\bm{y}$, while $G(\cdot)$ encodes the regularization, promoting prior knowledge about $\bm{x}$. Problem~\eqref{eq:opt} is typically solved using proximal splitting methods~\cite{parikh2014proximal}, especially when $G(\bm{x})$ is convex.

In proximal splitting, the proximal step associated with $G(\bm{x})$ can be interpreted as a denoising operation. This insight has led to replacing the proximal step with advanced denoisers, such as BM3D~\cite{dabov2006image,dabov2007image}, or learned neural network-based denoisers~\cite{zhang2017beyond, zhang2020plug, ryu2019plug}. More recently, generative models that learn image priors directly from data have been incorporated into the Plug-and-Play (PnP) framework~\cite{martin2024pnp, zhu2023denoising, song2023loss, graikos2022diffusion, hurault2022proximal, kamilov2017plug}. Although generative models can in principle provide $\log p(\bm{x})$~\cite{chen2018neural}, evaluating its gradient via backpropagation is computationally expensive. PnP methods circumvent this issue by directly substituting a denoiser induced by a generative model for the proximal operator. Such approaches have achieved state-of-the-art results in inverse problems using powerful generative priors, including diffusion models~\cite{chung2022diffusion, chung2022improving, song2022pseudoinverse, kawar2022denoising} and flow matching models~\cite{pokle2024training, zhang2024flow, martin2024pnp}.

However, prior PnP methods have predominantly focused on inverse problems with Gaussian noise, where the data fidelity term $F(\cdot)$ is typically the squared $\ell_2$ loss. While effective under Gaussian noise, this loss function is suboptimal for non-Gaussian noise models. For example, Poisson noise is the more appropriate modeling for the shot noise of photons in the observation. Salt-and-pepper noise, also known as impulse noise, is another form of noise in digital images. Although one can still use squared $\ell_2$ loss, the mismatch in noise modeling often leads to subpar restoration quality.

In this work, we extend PnP methods based on flow matching generative models to handle non-Gaussian noise by generalizing the data fidelity term $F(\cdot)$ in~\eqref{eq:opt}. Specifically, we consider proximity-friendly fidelity functions such as $\ell_1$ and $\ell_2$ norms, which are better suited for modeling Poisson and impulse noise, respectively. For sparse impulse noise, $\ell_1$-norm fidelity is widely adopted due to its robustness to outliers. While the Poisson likelihood leads naturally to a specific form of $F$, in practice, $\ell_2$ norm fidelity is often preferred for its numerical stability. Our approach unifies various noise models within a general PnP framework, encompassing squared $\ell_2$ loss as a special case.

The main contributions of this papers are as follows:
\begin{itemize}
\item We propose a generalized PnP framework using the primal-dual hybrid gradient (PDHG) method to solve~\eqref{eq:opt} with non-Gaussian noise, by accommodating flexible fidelity terms $F(\cdot)$.
\item The proximal step for $G(\cdot)$ is implemented using an implicit denoiser derived from flow matching generative models.
\item We seamlessly integrate flow matching generative models into the classical PDHG method, achieving an efficient and implementation-friendly solution.
\item Extensive experiments on two benchmark datasets across diverse tasks, including denoising, deblurring, inpainting, and super-resolution, demonstrate the superiority of our method under non-Gaussian noise, outperforming existing PnP methods designed specifically for Gaussian noise. 
\end{itemize}

\section{Background}

The performance of imaging inverse problems has significantly improved with the advent of generative models that learn expressive image priors. Among these, diffusion models~\cite{song2020score, ho2020denoising} and flow matching models~\cite{lipmanflow, tongimproving} have attracted considerable attention due to their superior generative capabilities compared to earlier approaches. These models have demonstrated remarkable potential for image restoration when used as priors within optimization frameworks.

In this section, we focus on flow matching generative models, emphasizing their simulation-free training and their application as denoisers. These denoisers play a critical role in the Plug-and-Play (PnP) approach for solving the regularized optimization problem in~\eqref{eq:opt}.

\subsection{Flow matching generative model}

Flow matching aims to learn a continuous transformation that maps samples from a simple distribution $p(\bm{x}_0)$ (\eg, Gaussian) to samples from a data distribution $p(\bm{x}_1)$. Let $\mathcal{P}(\mathbb{R}^n)$ denote the space of probability measures over $\mathbb{R}^n$, and consider two probability measures $P_0$ and $P_1$ representing the latent and data distributions, respectively. Define $\Gamma(P_0, P_1)$ as the set of couplings $\pi \in \mathcal{P}(\mathbb{R}^n \times \mathbb{R}^n)$ with marginals $P_0$ and $P_1$.

Flow matching constructs a continuous probability path $t \mapsto P_t$ for $t \in [0,1]$, interpolating between $P_0$ and $P_1$. Given a coupling $(\bm{x}_0, \bm{x}_1)$, one defines the interpolated point $\bm{x}_t = (1 - t)\bm{x}_0 + t\bm{x}_1$, which induces a trajectory governed by the ordinary differential equation (ODE):
\begin{equation}
  \label{eq:flow}
  \frac{d\bm{x}_t}{dt} = \bm{v}(\bm{x}_t,t),
\end{equation}
where $\bm{v}(\bm{x}_t, t)$ is the velocity field. The goal is to learn this velocity field from data. To this end, a neural network $\bm{v}_{\theta}(\bm{x}_t, t)$ parameterized by $\theta$ is used to approximate $\bm{v}(\bm{x}_t, t)$ such that the evolution of $P_t$ satisfies the \emph{continuity equation}:
\begin{equation}
  \partial_tP_t +\nabla\cdot (P_t\bm{v}(\bm{x}_t,t)) = 0.
\end{equation}
To enable the continuous transition from a simple density $p(\bm{x}_0)$ to a data density $p(\bm{x}_1)$, the core of flow matching is to define the velocity field $\bm{v}(\bm{x}_t,t)$ in~\eqref{eq:flow}. 

Similar to normalizing flows~\cite{kingma2018glow}, the flow matching model can be trained via maximum likelihood estimation:
\begin{equation}
  \mathcal{L}_{\text{ML}}(\theta) = \mathbb{E}_{\bm{x}_1\sim p(\bm{x}_1)} [\log p(\bm{x}_1)],
\end{equation}
where the expectation is taken over the dataset. The log-likelihood of $p_1(\bm{x})$ can be computed using the differential equation of evolution of the log-likelihood:
\begin{equation}
  \label{eq:p_e}
  \frac{\partial \log p(\bm{x}_t)}{\partial t} = -\trace\left(\frac{\bm{v}_{\theta}(\bm{x}_t,t)}{\partial\bm{x}_t}\right).
\end{equation}

One can additionally obtain the likelihood of the final $\log p(\bm{x}_1)$ via integrating~\eqref{eq:p_e}:
\begin{equation}
  \log p(\bm{x}_1) = \log p(\bm{x}_0) -\int_0^1\trace\left(\frac{\bm{v}_{\theta}(\bm{x}_s,s)}{\partial\bm{x}_s}\right)ds.
  \end{equation}
However, evaluating this integral is computationally intensive. To avoid this, simulation-free training methods have been proposed.

 The conditional flow matching offers such a simulation-free training approach~\cite{lipmanflow}. Its learning objective is
\begin{equation}
  \label{eq:cfm}
  \mathcal{L}_{\text{CFM}}(\theta)= \mathbb{E}_t[\mathbb{E}_{(\bm{x}_0,\bm{x}_1)\sim \pi} \norm{\bm{v}_{\theta}(\bm{x}_t,t)-(\bm{x}_1-\bm{x}_0)}_2^2],
\end{equation}
where $\pi$ is a coupling between $p(\bm{x}_0)$ and $p(\bm{x}_1)$. The minimization of loss~\eqref{eq:cfm} is equivalent to the minimization of the direct flow matching loss
\begin{equation}
  \label{eq:fm}
  \mathcal{L}_{\text{FM}}(\theta):= \mathbb{E}_t[\mathbb{E}_{\bm{x}_t\sim P_t} \norm{\bm{v}_{\theta}(\bm{x}_t,t)-\bm{v}(\bm{x}_t,t)}_2^2].
\end{equation}
The choice of coupling $\pi$ critically affects both training efficiency and generative quality. Optimal coupling can be obtained via standard optimal transport solvers~\cite{tongimproving}. Alternatively, rectified flow~\cite{liuflow} refines transport maps by iteratively straightening ODE trajectories, improving convergence and model quality.

\subsection{Denoiser induced by flow model}

Let $\bm{x}$ be a sample from the random variable $X_t := (1 - t)X_0 + tX_1$, where $X_0 \sim P_0$ and $X_1 \sim P_1$ respectively. For a fixed time $t\in [0,1]$, the minimizer $\bm{v}_t^*$ of the CFM loss~\eqref{eq:cfm} over all admissible vector fields is given by:
\begin{equation}
  \bm{v}_t^*(x) = \mathbb{E}[X_1-X_0|X_t=\bm{x}].
\end{equation}
Assume we are in the ideal case where $\bm{v}_t^{\theta}=\bm{v}_t^*$. Then it follows that, for any $\bm{x}$ and $t\in [0,1]$, the denoising operator
\begin{equation}
  D_t(\bm{x}) = \bm{x} + (1-t)\bm{v}_t^*(\bm{x})
\end{equation}
satisfies that $D_t(\bm{x})=\mathbb{E}[X_1|X_t=x]$, \ie, the conditional mean estimator of the clean image. It implies that one has a natural denoiser operator induced by a flow matching generative model. Thus, flow matching models naturally induce denoisers that can be directly integrated into proximal splitting methods for solving the MAP problem in~\eqref{eq:opt}. This framework enables a training-free, zero-shot solution for imaging inverse problems: once trained, the generative model can be applied to a wide range of inverse tasks without retraining or fine-tuning. Compared to methods that directly learn conditional score functions for specific tasks~\cite{whang2022deblurring}, our approach provides a universal and versatile framework for solving both linear and nonlinear inverse problems using pre-trained, off-the-shelf generative models. There is no need to fine-tune the model.

\section{Related works}

\subsection{Measurement-guided sampling with diffusion model}

Prior to the development of flow matching models, diffusion models emerged as a powerful class of generative models and have been widely adopted for image restoration tasks, including both linear and nonlinear inverse problems. The central idea in these approaches is to guide the sampling process using measurement information, typically through a gradient descent step integrated into the unconditional diffusion sampling scheme.

Various methods differ in how they incorporate measurement guidance via fidelity losses. \citet{chung2022diffusion} introduced a rough fidelity loss based on known likelihoods and Tweedie's formula. However, this approximation introduces a significant bias, negatively impacting restoration quality. To address this, \citet{song2022pseudoinverse} proposed $\Pi$GDM, which improves fidelity modeling for Gaussian noise. Nonetheless, $\Pi$GDM is tailored exclusively to Gaussian noise scenarios and does not generalize well to non-Gaussian or nonlinear problems.

Other notable methods include DDRM~\cite{kawar2022denoising} and DDNM/DDNM+~\cite{wang2022zero}, which leverage the singular value decomposition (SVD) of the degradation matrix $\bm{A}$ for spectral inpainting. However, SVD may be computationally infeasible or unavailable in many practical applications.

\subsection{Measurement-guided linear inversers via flows}

Building on the efficiency of flow matching models, several works have proposed measurement-guided sampling techniques for image restoration. In particular, \citet{pokle2024training} introduced a training-free flow-based method that incorporates a theoretically justified weighting scheme into the sampling process. This method modifies the standard flow model's unconditional sampling by introducing a gradient descent step to refine predictions using measurement information. Like diffusion-based counterparts, these methods aim to solve linear inverse problems but with improved efficiency and interpretability. 

\subsection{Plug-and-play (PnP) methods}

The Plug-and-Play (PnP) framework integrates powerful denoisers into iterative optimization algorithms for image restoration. \citet{zhu2023denoising} introduced PnP-HQS (also known as DiffPIR), which incorporates a denoiser derived from a pre-trained diffusion model into a half-quadratic splitting (HQS) framework. The algorithm proceeds along a restoration trajectory:
\begin{equation*}
  \bm{x}_T\to \cdots\to \bm{x}_t\to \hat{\bm{x}}_0(\bm{x}_t)\to \hat{\bm{x}}_0'(\bm{x}_t)\to \bm{x}_{t-1}\to\cdots\bm{x}_0.
\end{equation*}
 At each step, given $\bm{x}_t$, the current estimate $\bm{x}_t$ is denoised to obtain $\hat{\bm{x}}_0(\bm{x}_t)$. A measurement-guided update is then computed by solving the proximal problem:
\begin{equation*}
  \label{eq:diff}
  \hat{\bm{x}}_0'(\bm{x}_t)=\argmin_{\bm{x}}\quad \frac{1}{2}\norm{\bm{A}\bm{x}-\bm{y}}_2^2 + \frac{\lambda\sigma_t^2}{\alpha_t^2}\norm{\bm{x}-\mathbb{E}[\bm{x}_0|\bm{x}_t]}_2^2.
\end{equation*}
In a related work, \citet{10747553} employed a generative model to initialize a classical total variation (TV) regularized optimization, solved via the PDHG method, to address limited-angle CT reconstruction.

For flow matching models, \citet{martin2024pnp} proposed PnP-FBS, which integrates flow matching into a forward-backward splitting (FBS) framework. Their algorithm alternates between gradient descent on the data fidelity term and a denoising step using the flow model. Similar to PnP-HQS, a re-noising step is applied before denoising to match the noise level. However, both PnP-HQS and PnP-FBS consider only smooth squared $\ell_2$ fidelity terms and are limited to Gaussian noise settings.

\subsection{Differentiation through flow ODE}

The flow matching model defines a continuous transformation from a latent variable to a data sample via an ODE. Leveraging this structure, \citet{ben2024d} formulates D-Flow via an ODE-constrained optimization problem, wherein gradients with respect to the latent variable are computed using adjoint methods. Although this approach allows for exact computation of the negative log-likelihood, it incurs high computational cost due to integration. To alleviate this, Flow-Priors~\cite{zhang2024flow} introduced a sequential approximation strategy that sidesteps the need for costly integrals, enabling efficient MAP estimation with flow models.

\section{Proposed method}

While existing training-free Plug-and-Play (PnP) methods utilizing flow-based generative models have shown promising results for imaging inverse problems under Gaussian noise, their extension to non-Gaussian noise settings remains underexplored. Previous generative PnP methods, including PnP-HQS and PnP-FBS, primarily address Gaussian noise, resulting in data fidelity terms modeled via the squared $\ell_2$ loss. However, in practice, non-Gaussian noise types such as Poisson and impulse noise are common, for which alternative loss functions—specifically the $\ell_1$ or $\ell_2$ norms—often outperform the squared $\ell_2$ loss due to better noise modeling.

To address this gap, we propose a generalized PnP framework that integrates flow-based generative priors and supports both $\ell_1$ and $\ell_2$ norm-based data fidelity terms. Our key innovation lies in leveraging the Primal-Dual Hybrid Gradient (PDHG) method to efficiently handle non-smooth fidelity losses such as $\ell_1$ and $\ell_2$ norms, whose proximal operators can be computed in closed form.

\subsection{Primal dual hybrid gradient (PDHG)}

The PDHG method~\cite{chambolle2011first,esser2010general,o2020equivalence} is an effective optimization method designed to solve convex composite problems of the form in~\eqref{eq:opt}. The method relies on alternating updates involving the proximal operators of $G$ and the convex conjugate of $F$, denoted $F^*$. A more detailed discussion on proximal operators and conjugate functions is provided in Appendix\ref{sec:ad}. 

Given initializations $\bm{x}^0$ and $\bm{z}^0$, the PDHG updates proceed as follows:
\begin{equation}
  \label{eq:pdhg}
  \begin{aligned}
    \overline{\bm{x}}^k &= \prox_{\tau G}(\bm{x}^{k-1}-\tau \bm{A}^T\bm{z}^{k-1})\\
      \overline{\bm{z}}^k &= \prox_{\sigma F^*}(\bm{z}^{k-1}+\sigma \bm{A}(2\overline{\bm{x}}^k-\bm{x}^{k-1}))\\
      \bm{x}^k &= \bm{x}^{k-1} +\rho_k(\overline{\bm{x}}^k-\bm{x}^{k-1})\\
      \bm{z}^k &= \bm{z}^{k-1} + \rho_k(\overline{\bm{z}}^k-\bm{z}^{k-1}),
  \end{aligned}
\end{equation}
where $\tau > 0$ and $\sigma > 0$ are step sizes satisfying $\sigma\tau\norm{\bm{A}}^2\leq 1$, where $\norm{\bm{A}}$ denotes the spectral norm of $\bm{A}$, and $\rho_k \in (0,2)$ is a relaxation parameter. In our implementation, we set $\rho_k \equiv 1$ for all $k$.

The generative model induces a time-dependent denoiser $D_t(\cdot)$, which we use to approximate the proximal operator of the regularization term $G(\cdot)$. We take PDHG method~\eqref{eq:pdhg} to solve~\eqref{eq:opt}, and $\rho_k\equiv 1$ in PDHG. The remained thing is to compute the $G$-relating proximal operator, $F^*$-relating proximal operator and determine the stepsize. In the following subsections, we detail the implementation of the proximal operators for $G$ and $F^*$, as well as the selection of step sizes for different noise models.

\subsection{$G$-relating proximal operator as denoising step}

Given initializations $\bm{x}_0,\bm{z}_0$, for per iteration cycle, PDHG algorithm first computes
\begin{equation}
  \bm{x}_k = \prox_{\tau G}(\bm{x}_{k-1}-\tau \bm{A}^T\bm{z}_{k-1}).
\end{equation}
The key is to compute the $G$-relating proximal operator
\begin{equation}
  \begin{aligned}
    \bm{v} &= \bm{x}_{k-1}-\tau \bm{A}^T\bm{z}_{k-1}\\
    \bm{x}_k&=\argmin_{\bm{z}}\quad \{\tau G(\bm{z}) + \frac{1}{2}\norm{\bm{z}-\bm{v}}_2^2\}.
  \end{aligned}
\end{equation}
Now it's time to build the connection between the proximal operator and the denoiser from generative models. For the flow matching model, assume that we want to resolve the noiseless $\bm{x}_k$ from $\bm{v}$ with a time-dependent noise level related to $\tau=\tau_k = t_k^2$. If we assume $G(\bm{z})\simeq -\log p_{\text{data}}(\bm{z})$ and the conditional density $p(\bm{v}|\bm{z})\simeq \log p_t(\bm{v}|\bm{z})$, then the proximal operator acts as a denoiser $D_t(\bm{v})$ induced from the flow generative model. 

However, the input $\bm{v}$ does not necessarily have the noise level assumed by the denoiser $D_t(\cdot)$. To address this mismatch, we follow the reprojection technique widely used in the literature~\cite{zhu2023denoising,martin2024pnp}. Assuming $\bm{v}$ lies approximately on the data manifold (an idealization that works well in practice), we reproject it to the appropriate noise level by adding Gaussian noise.

In summary, the $G$-relating proximal step can be implemented as follows:
\begin{enumerate}
\item \textbf{Variable $\bm{z}$ update}: which performed $\bm{z} = \bm{x}_{k-1}-\tau \bm{A}^T\bm{z}_{k-1}$
\item \textbf{Reprojection step}: which added noise to match the assumed noise level. For flow generative models, the renoising steps are as follows
  \begin{equation*}
      \bm{x}_k = (1-t_k)\cdot\bm{\epsilon}+t_k \cdot \bm{z}, \bm{\epsilon}\sim\mathcal{N}(0,\bm{I})
  \end{equation*}
\item \textbf{Denoising step}: which utilized the generative model to return a denoised version $\bm{x}_k=D_t(\bm{x}_k)$. 
\end{enumerate}

\subsection{$F^*$-relating proximal operator for data fidelity}

The next step of PDHG is related to the data fidelity term $F$. By Moreau’s identity~\eqref{eq:moreau}:
\begin{equation}
  \label{eq:moreau}
  \prox_{\sigma f^*}(\bm{y}) = \bm{y} - \sigma\prox_{\sigma^{-1}f}(\sigma^{-1}\bm{y}),
\end{equation}
we are required to compute the proximal operator of function $F(\cdot)$. For Gaussian noise measurement, the squared $\ell_2$ norm is the standard choice. For sparse noise, the $\ell_1$ norm is more suitable over the $\ell_2$ norm-based data fidelity. For Poisson noise, $\ell_2$ norm-based loss outperforms the maximum likelihood estimation based loss. For sparse impulse noise, $\ell_1$ norm is the standard choice. Each of these fidelity terms yields closed-form proximal operators, which are summarized in Table~\ref{tab:1}.

\subsection{Adaptive stepsize selection}

To guarantee the convergence of PDHG algorithm, the stepsizes $\tau,\sigma$ shall satisfy $\tau\sigma\norm{\bm{A}}^2\leq 1$. As previously mentioned, the stepsize $\tau$ shall configured with the assumed noise level. For flow matching or rectified flow model, one can set $\tau_k=\gamma (t_k)^{s}$ ($s>0$). For the two flow models, when $s=2$ and $\gamma=1$, it degenerates to the assumptions we used. The dual step size is set as $\sigma_k=1/\tau_k$. The choice of $s$ affects the final restoration performance, we tuned it in experiments. Note that other options of the stepsize can be tried. Despite possible variations, the above step size schedule demonstrated strong empirical performance across tasks. The algorithm flowchart is detailed in Alg.~\ref{alg:2}.

\begin{table}[!htp]
  \caption{Three common losses and their proximal operators.\label{tab:1}}
  \centering
  \begin{tabular}{c|ccccc}
    \toprule
    Loss $\lambda F(\bm{x})$ & Proximal operator $\prox_{\lambda F}(\bm{x})$\\
    \midrule
    $\frac{\lambda}{2}\norm{\bm{x}-\bm{y}}_2^2$ & $\frac{\bm{x}+\lambda\bm{y}}{1+\lambda}$
    \\
    $\lambda\norm{\bm{x}-\bm{y}}_1$ & $\bm{y}+\text{sign}(\bm{x}-\bm{y})\max(|\bm{x}-\bm{y}|-\lambda,0)$
    \\
    $\lambda\norm{\bm{x}-\bm{y}}_2$ & $\bm{y}+\left(1-\frac{\lambda}{\max\{\norm{\bm{x}-\bm{y}}_2,\lambda\}}\right)(\bm{x}-\bm{y})$\\
        \bottomrule
  \end{tabular}\vspace*{-10pt}
\end{table}

%\vspace*{-5pt}
\begin{algorithm}[!htbp]
\caption{PHGD method with PnP flow generative prior.\label{alg:2}}
\begin{algorithmic}[1]
   \REQUIRE {Iterations $T$, Loss $F$, Denoiser induced by generative model $D_k(\cdot)$, stepsize $\eta_k=\eta\tau_k$. }
 \ENSURE {Estimated  image $\bm{x}_0$.}
  \STATE Set $\bm{x}_0\sim \mathcal{N}(0,\bm{I})$ and $\bm{z}_0= \bm{0}$
  \FOR{$k = 1:1:T$}
  \STATE $\bm{z}$ update: $\bm{z}=\bm{x}_{k-1}-\eta\tau_{k-1}\bm{A}^T\bm{z}_{k-1}$
  \STATE Reprojection step:
 \begin{equation*}
      \bm{x}_k = (1-t_k)\cdot\bm{\epsilon}+t_k \cdot\bm{z}, \bm{\epsilon}\sim\mathcal{N}(0,\bm{I}) 
  \end{equation*}
  \STATE Denoising step: $\bm{x}_k = D_k(\bm{x}_k)$
   \STATE $F^*$-relating step: $\bm{z}_k=\text{prox}_{\tau_{k-1}^{-1}F^*}(\bm{z}_{k-1}+\tau_{k-1}^{-1}\bm{A}(2\bm{x}_{k}-\bm{x}_{k-1}))$.
\ENDFOR
\end{algorithmic}
\end{algorithm}\vspace*{-10pt}

\begin{table*}[!htp]
  \centering
  \small
    \vspace*{-10pt}
    \fontsize{9}{12}\selectfont 
    \caption{Quantitative results of different methods for four tasks with Poisson/salt-and-pepper noise. \label{tab:poisson}}
    \scalebox{0.9}{\begin{tabular}{c@{\hspace{1pt}}c@{\hspace{1pt}}c@{\hspace{1pt}}c@{\hspace{1pt}}c@{\hspace{1pt}}c@{\hspace{1pt}}c@{\hspace{1pt}}c@{\hspace{1pt}}c@{\hspace{1pt}}c@{\hspace{1pt}}c@{\hspace{1pt}}c@{\hspace{1pt}}c@{\hspace{1pt}}c@{\hspace{1pt}}c@{\hspace{1pt}}c@{\hspace{1pt}}c@{\hspace{1pt}}c@{\hspace{1pt}}c@{\hspace{1pt}}c@{\hspace{1pt}}c@{\hspace{1pt}}c@{\hspace{1pt}}c@{\hspace{1pt}}c@{\hspace{1pt}}c@{\hspace{1pt}}c@{\hspace{1pt}}c@{\hspace{1pt}}c@{\hspace{1pt}}c@{\hspace{1pt}}c@{\hspace{1pt}}c@{\hspace{0pt}}cc@{\hspace{1pt}}c@{\hspace{1pt}}c@{\hspace{0pt}}cc@{\hspace{1pt}}c@{\hspace{1pt}}c@{\hspace{0pt}}}
  & &  \multicolumn{11}{c}{\textbf{(a) Poisson}} && \multicolumn{11}{c}{\textbf{(b) salt-and-pepper }} \\
 \toprule
 \multirow{2}{*}{\STAB{\rotatebox[origin=c]{90}{Data}}} & \multirow{2}{*}{Method}  &   \multicolumn{2}{c}{\textbf{Denoising}}  && \multicolumn{2}{c}{\textbf{Deblur}} && \multicolumn{2}{c}{\textbf{SR}}  && \multicolumn{2}{c}{\textbf{Box Inp}} && \multicolumn{2}{c}{\textbf{Denoising}}  && \multicolumn{2}{c}{\textbf{Deblur}} && \multicolumn{2}{c}{\textbf{SR}}  && \multicolumn{2}{c}{\textbf{Box Inp}}\\
        \cline{3-4} \cline{6-7} \cline{9-10} \cline{12-13} \cline{15-16} \cline{18-19} \cline{21-22} \cline{24-25}
     & & PSNR  & SSIM && PSNR  & SSIM  && PSNR  & SSIM && PSNR  & SSIM && PSNR  & SSIM && PSNR  & SSIM  && PSNR  & SSIM && PSNR  & SSIM\\
   \toprule
  \multirow{4}{*}{\STAB{\rotatebox[origin=c]{90}{FFHQ}}} & Flow-Priors & 25.07 & 0.654 &  & 18.44 & 0.369 &  & 20.89 & 0.442 &  & 20.99 & 0.461 && 26.83 & 0.677 &  & 18.54 & 0.334 &  & 22.17 & 0.621 &  & 22.86 & 0.638 \\
&OT-ODE      & 23.66 & 0.596 &  & 22.73 & 0.567 &  & 21.64 & 0.518 &  & 22.48 & 0.566 && 22.32 & 0.478 &  & 22.79 & 0.549 &  & 18.98 & 0.372 &  & 22.79 & 0.575\\
&PnP-HQS     & 23.90  & 0.729 &  & 23.25 & 0.703 &  & 18.76 & 0.578 &  & 22.97 & 0.709 && 23.95 & 0.633 &  & 22.66 & 0.591 &  & 23.31 & 0.696 &  & 22.34 & 0.548\\
&PnP-FBS     & 29.72 & 0.901 &  & 28.45 & 0.855 &  & 23.51 & 0.590  &  & 27.25 & 0.877 && 26.27 & 0.785 &  & 25.57 & 0.786 &  & 21.92 & 0.693 &  & 24.90  & 0.749\\
&Ours        & \textbf{33.98} & \textbf{0.902} &  & \textbf{32.87} & \textbf{0.927} &  & \textbf{29.96} & \textbf{0.836} &  & \textbf{29.48} & \textbf{0.885} && \textbf{40.27} & \textbf{0.974} &  & \textbf{37.67} & \textbf{0.974} &  & \textbf{33.12} & \textbf{0.948} &  & \textbf{30.57} & \textbf{0.948}\\
   \bottomrule
\toprule
 \multirow{4}{*}{\STAB{\rotatebox[origin=c]{90}{AFHQ-Cat}}} &  Flow-Priors & 24.65 & 0.670  &  & 20.48 & 0.482 &  & 18.80  & 0.399 &  & 20.01 & 0.390 && 26.98 & 0.719 &  & 21.37 & 0.545 &  & 14.14 & 0.336 &  & 22.79 & 0.551  \\
&OT-ODE      & 23.77 & 0.540  &  & 21.31 & 0.422 &  & 18.65 & 0.353 &  & 20.83 & 0.496 && 22.79 & 0.457 &  & 22.09 & 0.453 &  & 20.06 & 0.389 &  & 21.30  & 0.535 \\
&PnP-HQS     & 24.57 & 0.668 &  & 23.03 & 0.608 &  & 17.02 & 0.470  &  & 22.81 & 0.645 && 25.66 & 0.693 &  & 21.86 & 0.595 &  & 18.86 & 0.521 &  & 22.46 & 0.565 \\
&PnP-FBS     & 29.65 & 0.854 &  & 25.57 & 0.695 &  & 23.05 & 0.558 &  & 25.42 & 0.815 && 27.21 & 0.791 &  & 24.14 & 0.648 &  & 19.24 & 0.530  &  & 24.44 & 0.744 \\
&Ours        & \textbf{34.82} & \textbf{0.935} &  & \textbf{26.28} & \textbf{0.709} &  & \textbf{26.26} & \textbf{0.747} &  & \textbf{26.25} & \textbf{0.889} && \textbf{37.94} & \textbf{0.962} &  & \textbf{28.05} & \textbf{0.791} &  & \textbf{24.70}  & \textbf{0.730}  &  & \textbf{25.43} & \textbf{0.908}\\
\bottomrule
 \end{tabular}}\vspace*{-10pt}        
  \end{table*}

\subsection{Extension to score-based diffusion model}

Although diffusion models are typically formulated using stochastic differential equations (SDEs), they are mathematically equivalent to ordinary differential equations (ODEs) governing the sampling trajectories. This connection enables our proposed PDHG-based plug-and-play (PnP) method to be applied to diffusion models as well, utilizing the denoiser implicitly defined by the diffusion process.

Given the Gaussian conditional density $p(\bm{x}_t|\bm{x}_0)\sim\mathcal{N}(\bm{x}_0;\alpha_t\bm{x}_0,\sigma_t^2\bm{I})$, the mean of $p(\bm{x}_0|\bm{x}_t)$ is given by the Tweedie's formula:
\begin{equation}
  \mathbb{E}[\bm{x}_0|\bm{x}_t] = \frac{\bm{x}_t+\sigma_t^2\nabla_{\bm{x}_t}\log p(\bm{x}_t)}{\alpha_t},
\end{equation}
where the score function $\nabla_{\bm{x}_t}\log p(\bm{x}_t)$ is replaced by the learned generative model in practice.

Note that, the mean estimation of $p(\bm{x}_0|\bm{x}_t)$ just defines a denoiser $D_t(\bm{x}_t)$. For variance persevering configuration of diffusion model, \ie, DDPM, we learned the scaled score function, hence the denoiser is given by
\begin{equation}
  D_t(\bm{x}_t) = \frac{\bm{x}_t-\sigma_t\bm{\epsilon}_{\theta}(\bm{x}_t,t)}{\alpha_t}.
\end{equation}
For variance exploding model, the corresponding denoiser can be achieved as well.

\section{Numerical experiments}

Our proposed PnP method is model-agnostic and compatible with any ODE-based generative model, including the flow matching model~\cite{lipmanflow}, diffusion model~\cite{ho2020denoising}, and rectified flow model~\cite{liuflow}. We have evaluated the algorithm across several applications against these three generative models. For page limitation, we provide the results for flow matching model in the following content. We mainly show the advantages of our method over the existing works for non-Gaussian noise. Thus, we considered Poisson noise and salt-and-pepper noise in our experiments, as well as Gaussian noise as a baseline for comparison.

\begin{figure*}[!htp]
    \centering %\vspace*{50pt}
    \begin{tabular}{c@{\hspace*{3pt}}c@{\hspace*{1pt}}c@{\hspace*{1pt}}c@{\hspace*{1pt}}c@{\hspace*{1pt}}c@{\hspace*{1pt}}c@{\hspace*{1pt}}c@{\hspace*{1pt}}c@{\hspace*{1pt}}c@{\hspace*{2pt}}c@{\hspace*{2pt}}c@{\hspace*{2pt}}c@{\hspace*{1pt}}}
  \rotatebox[origin=c]{90}{Denoising} &\raisebox{-0.5\height}{\includegraphics[width = 0.13\textwidth]{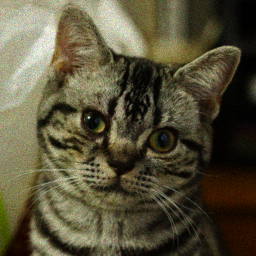}}
  & \raisebox{-0.5\height}{\includegraphics[width = 0.13\textwidth]{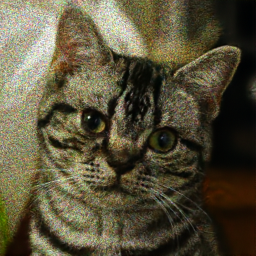}}
  & \raisebox{-0.5\height}{\includegraphics[width = 0.13\textwidth]{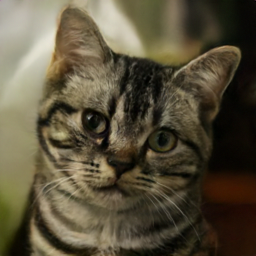}}
  & \raisebox{-0.5\height}{\includegraphics[width = 0.13\textwidth]{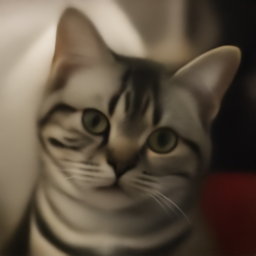}}
  & \raisebox{-0.5\height}{\includegraphics[width = 0.13\textwidth]{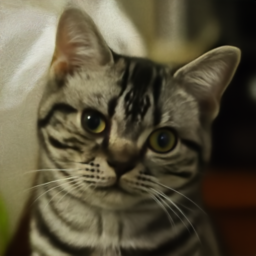}}
  & \raisebox{-0.5\height}{\includegraphics[width = 0.13\textwidth]{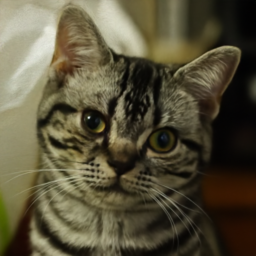}}
  & \raisebox{-0.5\height}{\includegraphics[width = 0.13\textwidth]{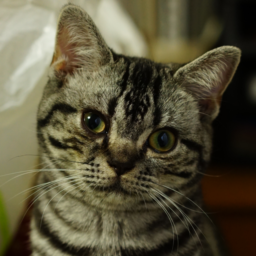}}\\[-2pt]
  \rotatebox[origin=c]{90}{Deblur} &\raisebox{-0.5\height}{\includegraphics[width = 0.13\textwidth]{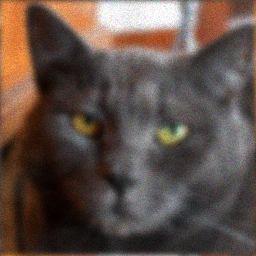}}
  & \raisebox{-0.5\height}{\includegraphics[width = 0.13\textwidth]{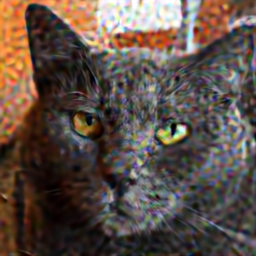}}
  & \raisebox{-0.5\height}{\includegraphics[width = 0.13\textwidth]{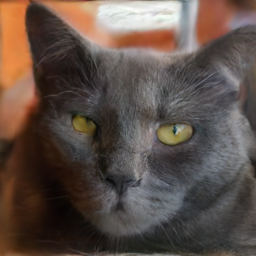}}
  & \raisebox{-0.5\height}{\includegraphics[width = 0.13\textwidth]{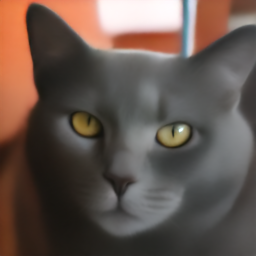}}
  & \raisebox{-0.5\height}{\includegraphics[width = 0.13\textwidth]{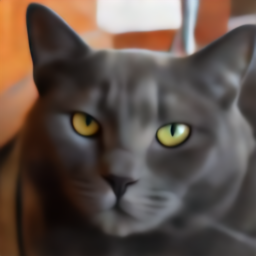}}
  & \raisebox{-0.5\height}{\includegraphics[width = 0.13\textwidth]{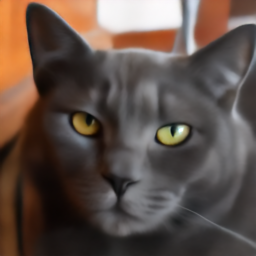}}
      & \raisebox{-0.5\height}{\includegraphics[width = 0.13\textwidth]{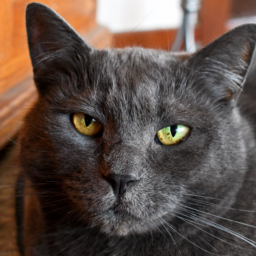}}\\[-2pt]
  \rotatebox[origin=c]{90}{SR} &\raisebox{-0.5\height}{\includegraphics[width = 0.13\textwidth]{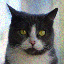}}
  & \raisebox{-0.5\height}{\includegraphics[width = 0.13\textwidth]{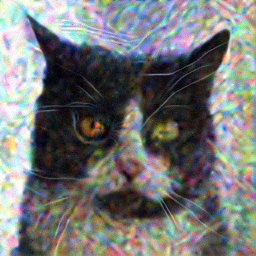}}
  & \raisebox{-0.5\height}{\includegraphics[width = 0.13\textwidth]{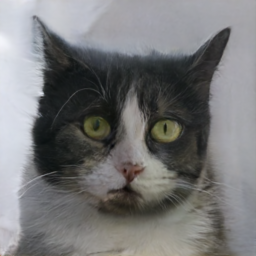}}
  & \raisebox{-0.5\height}{\includegraphics[width = 0.13\textwidth]{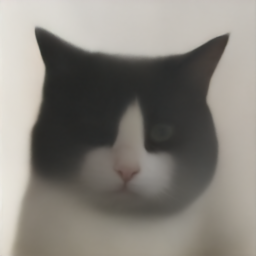}}
  & \raisebox{-0.5\height}{\includegraphics[width = 0.13\textwidth]{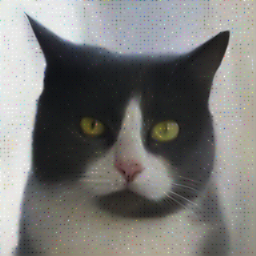}}
  & \raisebox{-0.5\height}{\includegraphics[width = 0.13\textwidth]{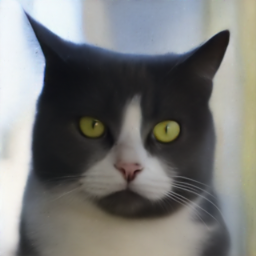}}
      & \raisebox{-0.5\height}{\includegraphics[width = 0.13\textwidth]{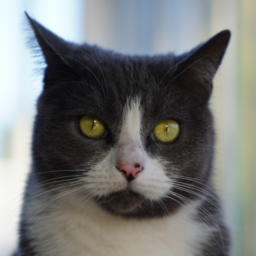}}\\[-2pt]
  \rotatebox[origin=c]{90}{Box inp} &\raisebox{-0.5\height}{\includegraphics[width = 0.13\textwidth]{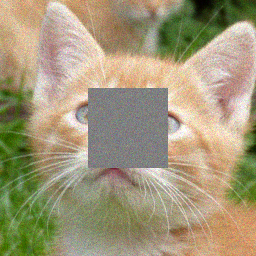}}
  & \raisebox{-0.5\height}{\includegraphics[width = 0.13\textwidth]{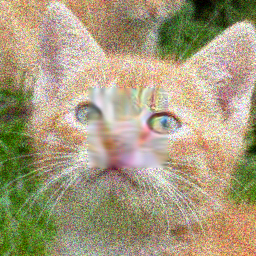}}
  & \raisebox{-0.5\height}{\includegraphics[width = 0.13\textwidth]{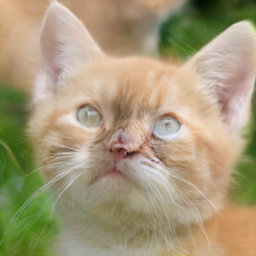}}
  & \raisebox{-0.5\height}{\includegraphics[width = 0.13\textwidth]{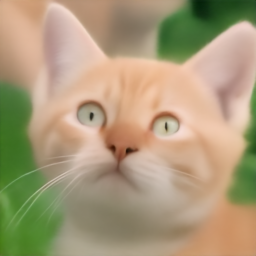}}
  & \raisebox{-0.5\height}{\includegraphics[width = 0.13\textwidth]{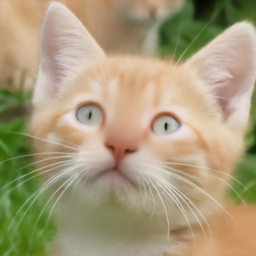}}
  & \raisebox{-0.5\height}{\includegraphics[width = 0.13\textwidth]{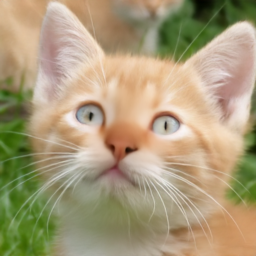}}
      & \raisebox{-0.5\height}{\includegraphics[width = 0.13\textwidth]{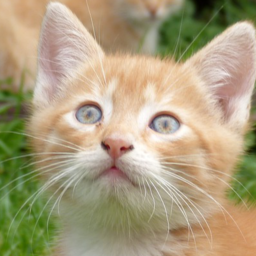}}\\
     &  \small{Input} & \small{Flow Priors} & \small{OT-ODE} &\small{PnP-HQS} & \small{PnP-Flow} & \small{Ours} & \small{GT}\\
  \end{tabular}%\vspace*{-8pt}
  \caption{\textbf{Visualization of different methods on four tasks with Poisson noise for AFHQ-Cat.}\label{fig:poisson} } \vspace*{-20pt}
  \end{figure*} 
  
  \begin{figure*}[!htp]
      \centering %\vspace*{50pt}
      \begin{tabular}{c@{\hspace*{3pt}}c@{\hspace*{1pt}}c@{\hspace*{1pt}}c@{\hspace*{1pt}}c@{\hspace*{1pt}}c@{\hspace*{1pt}}c@{\hspace*{1pt}}c@{\hspace*{1pt}}c@{\hspace*{1pt}}c@{\hspace*{2pt}}c@{\hspace*{2pt}}c@{\hspace*{2pt}}c@{\hspace*{1pt}}}
   \rotatebox[origin=c]{90}{Denoising} &\raisebox{-0.5\height}{\includegraphics[width = 0.13\textwidth]{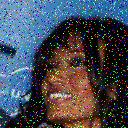}}
  & \raisebox{-0.5\height}{\includegraphics[width = 0.13\textwidth]{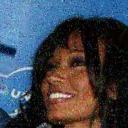}}
  & \raisebox{-0.5\height}{\includegraphics[width = 0.13\textwidth]{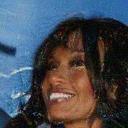}}
  & \raisebox{-0.5\height}{\includegraphics[width = 0.13\textwidth]{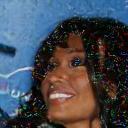}}
  & \raisebox{-0.5\height}{\includegraphics[width = 0.13\textwidth]{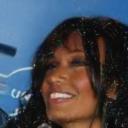}}
  & \raisebox{-0.5\height}{\includegraphics[width = 0.13\textwidth]{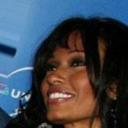}}
      & \raisebox{-0.5\height}{\includegraphics[width = 0.13\textwidth]{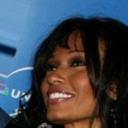}}\\[-2pt]
      \rotatebox[origin=c]{90}{Deblur} &\raisebox{-0.5\height}{\includegraphics[width = 0.13\textwidth]{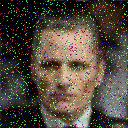}}
  & \raisebox{-0.5\height}{\includegraphics[width = 0.13\textwidth]{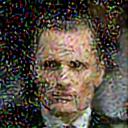}}
  & \raisebox{-0.5\height}{\includegraphics[width = 0.13\textwidth]{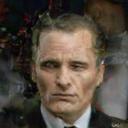}}
  & \raisebox{-0.5\height}{\includegraphics[width = 0.13\textwidth]{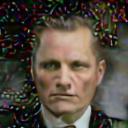}}
  & \raisebox{-0.5\height}{\includegraphics[width = 0.13\textwidth]{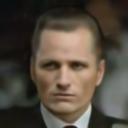}}
  & \raisebox{-0.5\height}{\includegraphics[width = 0.13\textwidth]{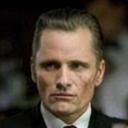}}
      & \raisebox{-0.5\height}{\includegraphics[width = 0.13\textwidth]{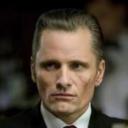}}\\[-2pt]
      \rotatebox[origin=c]{90}{SR} &\raisebox{-0.5\height}{\includegraphics[width = 0.13\textwidth]{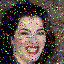}}
  & \raisebox{-0.5\height}{\includegraphics[width = 0.13\textwidth]{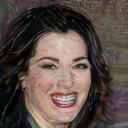}}
  & \raisebox{-0.5\height}{\includegraphics[width = 0.13\textwidth]{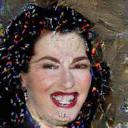}}
  & \raisebox{-0.5\height}{\includegraphics[width = 0.13\textwidth]{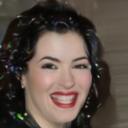}}
  & \raisebox{-0.5\height}{\includegraphics[width = 0.13\textwidth]{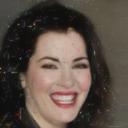}}
  & \raisebox{-0.5\height}{\includegraphics[width = 0.13\textwidth]{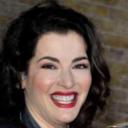}}
      & \raisebox{-0.5\height}{\includegraphics[width = 0.13\textwidth]{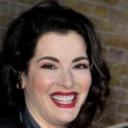}}\\[-2pt]
      \rotatebox[origin=c]{90}{Box inp} &\raisebox{-0.5\height}{\includegraphics[width = 0.13\textwidth]{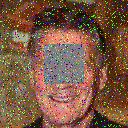}}
  & \raisebox{-0.5\height}{\includegraphics[width = 0.13\textwidth]{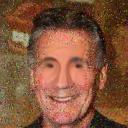}}
  & \raisebox{-0.5\height}{\includegraphics[width = 0.13\textwidth]{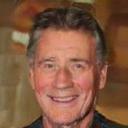}}
  & \raisebox{-0.5\height}{\includegraphics[width = 0.13\textwidth]{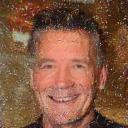}}
  & \raisebox{-0.5\height}{\includegraphics[width = 0.13\textwidth]{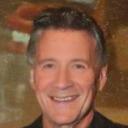}}
  & \raisebox{-0.5\height}{\includegraphics[width = 0.13\textwidth]{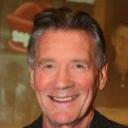}}
  & \raisebox{-0.5\height}{\includegraphics[width = 0.13\textwidth]{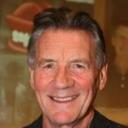}}\\
  &  \small{Input} & \small{Flow Priors} & \small{OT-ODE} &\small{PnP-HQS} & \small{PnP-Flow} & \small{Ours}  & \small{GT}\\
  \end{tabular}%\vspace*{-8pt}
    \caption{\textbf{Visualization of different methods on four tasks with impulse noise for CelebA.}\label{fig:sp}} \vspace*{-8pt}
  \end{figure*} %\vspace*{-18pt}

\subsection{Flow matching generative model}

\noindent{\bf Baselines}\ \ We benchmark our method against four PnP approaches with flow matching imaging prior, including OT-ODE~\footnote{With the unified framework, it is the flow matching based counterpart of diffusion posterior sampling method.}, Flow priors, PnP-FBS and PnP-HQS.
\\[2pt]
\noindent{\bf Benchmark datasets and publicly available models}\ \
For easy and fair comparison, we consider the datasets where there are publicly available models. For flow matching, we have the models for CelebA dataset with size $128\times 128$, and AFHQ-Cat, a subset of the Animal FacesHQ dataset focus on the cat class, with image size $256\times 256$. All images are normalized to the range $[-1,1]$. For CelebA, we use the standard training, validation and test splits. For AFHQ-Cat, we follow~\cite{martin2024pnp} to select random 32 images to create a validation set. For the two datasets, the pretrained model~\cite{martin2024pnp} is trained using the mini batch OT Flow Matching approach~\cite{tongimproving} for efficiency.
\\[2pt]
\noindent{\bf Evaluated imaging problems}\ \
The considered four image restoration problems are as follows. (a) Denoising; (b) deblurring using a $61\times 61$ Gaussian kernel; (c) supper-resolution with average pool downsampling operation ($2\times$ downsampling for image size $128\times 128$ and $4\times$ downsampling for image size $256\times 256$); (d) box-inpainting with a centered $s\times s$ mask ($s=40$ for image size $128\times 128$ and $s=80$ downsampling for image size $256\times 256$). 
\\[2pt]
\noindent{\bf Noise types}\ \
We simulate Poisson noise, salt-and-pepper noise and Gaussian noise for the clean observations as follows.
\begin{enumerate}
\item Poisson noise: which is simulated using $\bm{y} = \bm{z}/\alpha, \, \bm{z}\sim \text{Poisson}(\bm{A}\bm{x}_0\cdot\alpha)$,
where $\alpha$ denotes the noise level, we consider $\alpha=1$ in experiment. We set loss $F=\lambda\norm{\bm{A}\bm{x}-\bm{y}}_2$ ($\lambda=200$) for better performance over $\ell_1$ norm. 
\item Salt-and-pepper noise: which is a sparse noise in digital images. The noise consists of $10\%$ random pixels in image being set to black and white. The data fidelity term is $F=\lambda\norm{\bm{A}\bm{x}-\bm{y}}_1$ ($\lambda=25$).
\item Gaussian noise: Noise level $\sigma=0.2$ for denoising, $\sigma=0.05$ for other tasks (except $\sigma=0.01$ for random inpainting). The data fidelity term is $F=\frac{1}{2\sigma_y^2}\norm{\bm{A}\bm{x}-\bm{y}}_2^2$, where $\sigma_y$ is the noise level.
\end{enumerate}

\noindent{\bf Stepsize setting and iterations}\ \ In all experiments, our method sets a time-dependent stepsize $\tau_k$ and $\sigma_k=1/\tau_k$. For flow model, the stepsize $\tau_k$ is of the form $(1-t_k)^\alpha$ with $\alpha$ tuned for different tasks and datasets. In the case of Poisson and salt-and-pepper noise: we set $\alpha=0.8$ for all tasks.  In the case of Gaussian noise: $\alpha=0.8$ for denoising; $\alpha=0.01$ for deblurring; $\alpha=0.3$ for super-resolution and $\alpha=0.5$ for box-inpainting. We use 100 total sampling steps. All baseline hyperparameters are tuned following~\cite{martin2024pnp}.
\\[2pt]
\noindent{\bf Quantitative results}\ \
For quantitative comparison, we provide the reference-based PSNR and SSIM metrics. Tables~\ref{tab:poisson} (a) and (b) report PSNR and SSIM on CelebA and AFHQ-Cat for Poisson and impulse noise. For both datasets, our method beats other methods across all the four tasks and the two considered datasets by a large margin. The change of data consistency loss is crucial for the final image restoration. Excluding ours, other methods relies on the squared $\ell_2$ norm loss and we tried our best to tune the compared methods. The different performances of PnP-HQS and PnP-FBS are interesting, since they only differ from whether we solve the proximal problem exactly or just perform a gradient step. PnP-FBS is more robust to the mismatch between the noise type and the consistency loss. See Figures~\ref{fig:poisson} and~\ref{fig:sp} for the visualization of the different methods on the two datasets. Ours yields restoration with better details,  better data consistency, and fewer artifacts. See more visualization in the Appendix.

Table~\ref{tab:cele_ga} report the quantitative results for Gaussian noise. It can see that our method is comparable to PnP-FBS method and PnP-HQS method for the Gaussian noise scenario. The comparable result is reasonable since all the methods solve the same regularized optimization problem. Hence, the significant improvement is attributed to the exchange of the loss for non-Gaussian noise.

  \begin{table}[!htp]
  \centering
  \small
    %\vspace*{-10pt}
    \fontsize{9}{12}\selectfont 
  \caption{Quantitative results of different methods for different tasks on the two datasets with Gaussian noise. \label{tab:cele_ga}}
 \begin{tabular}{@{\hspace{0pt}}c@{\hspace{1pt}}c@{\hspace{1pt}}c@{\hspace{1pt}}c@{\hspace{1pt}}c@{\hspace{1pt}}c@{\hspace{1pt}}c@{\hspace{1pt}}c@{\hspace{1pt}}c@{\hspace{1pt}}c@{\hspace{1pt}}c@{\hspace{1pt}}c@{\hspace{1pt}}c@{\hspace{1pt}}c@{\hspace{1pt}}c@{\hspace{1pt}}c@{\hspace{1pt}}cc@{\hspace{1pt}}c@{\hspace{1pt}}c@{\hspace{0pt}}cc@{\hspace{1pt}}c@{\hspace{1pt}}c@{\hspace{0pt}}cc@{\hspace{1pt}}c@{\hspace{1pt}}c@{\hspace{0pt}}}
 \toprule
  CelebA  &   \multicolumn{2}{c}{\textbf{Denoising}}  && \multicolumn{2}{c}{\textbf{Deblur}} && \multicolumn{2}{c}{\textbf{SR}}  && \multicolumn{2}{c}{\textbf{Box Inp}} \\
        \cline{2-3} \cline{5-6} \cline{8-9} \cline{11-12}
     Method & PSNR  & SSIM && PSNR  & SSIM  && PSNR  & SSIM && PSNR  & SSIM \\
   \toprule
  Flow-Priors & 29.15 & 0.771 &  & 31.17 & 0.858 &  & 28.17 & 0.721 &  & 29.03 & 0.861 \\
OT-ODE      & 26.90  & 0.681 &  & 28.48 & 0.734 &  & 27.70  & 0.743 &  & 26.04 & 0.728 \\
PnP-HQS     & 31.91 & \textbf{0.912} &  & 20.55 & 0.617 &  & 22.68 & 0.692 &  & \textbf{29.84} & \textbf{0.944} \\
PnP-FBS     & \textbf{31.95} & 0.910  &  & \textbf{34.18} & \textbf{0.940}  &  & \textbf{30.90}  & \textbf{0.905} &  & 29.72 & 0.938 \\
Ours        & 31.40  & 0.904 &  & 33.90  & 0.935 &  & 30.37 & 0.898 &  & 29.20  & 0.931\\
   \bottomrule
\toprule
 AFHQ-Cat   &   \multicolumn{2}{c}{\textbf{Denoising}}  && \multicolumn{2}{c}{\textbf{Deblur}} && \multicolumn{2}{c}{\textbf{SR}}  && \multicolumn{2}{c}{\textbf{Box Inp}} \\
        \cline{2-3} \cline{5-6} \cline{8-9} \cline{11-12}
     Method & PSNR  & SSIM && PSNR  & SSIM  && PSNR  & SSIM && PSNR  & SSIM \\
   \toprule
  Flow-Priors &  29.86  & 0.779 &  &   24.51    & 0.682   &  &   22.29 & 0.499 &  &   26.48 & 0.820  \\
OT-ODE      & 26.75 & 0.635 &  & 24.71 & 0.552 &  & 23.88 & 0.568 &  & 22.81 & 0.677 \\
PnP-HQS     & \textbf{31.31} & \textbf{0.880}  &  & 21.66 & 0.573 &  & 19.18 & 0.518 &  & 26.67 & 0.917 \\
PnP-FBS     & 31.03 & 0.865 &  & \textbf{27.25} & \textbf{0.746} &  & \textbf{22.59} & \textbf{0.630}  &  & \textbf{26.83} & \textbf{0.904} \\
Ours        & 30.73 & 0.861 &  & 27.09 & 0.742 &  & 22.51 & 0.631 &  & 26.31 & 0.897 \\
\bottomrule
\end{tabular}
\end{table}

\begin{table}[!htp]
  \centering
  \small
  \vspace*{-8pt}
  
    \fontsize{9}{12}\selectfont 
  \caption{Comparison of $\ell_1$/$\ell_2$ losses for AFHQ-Cat. \label{tab:afhq}}%\vspace*{-8pt}
 \scalebox{0.9}{\begin{tabular}{@{\hspace{0pt}}c@{\hspace{1pt}}c@{\hspace{1pt}}c@{\hspace{1pt}}c@{\hspace{1pt}}c@{\hspace{1pt}}c@{\hspace{1pt}}c@{\hspace{1pt}}c@{\hspace{1pt}}c@{\hspace{1pt}}c@{\hspace{1pt}}c@{\hspace{1pt}}c@{\hspace{1pt}}c@{\hspace{1pt}}c@{\hspace{1pt}}c@{\hspace{1pt}}c@{\hspace{1pt}}cc@{\hspace{1pt}}c@{\hspace{1pt}}c@{\hspace{0pt}}cc@{\hspace{1pt}}c@{\hspace{1pt}}c@{\hspace{0pt}}cc@{\hspace{1pt}}c@{\hspace{1pt}}c@{\hspace{0pt}}}
 \toprule
  AFHQ  & &   \multicolumn{2}{c}{\textbf{Denoising}}  && \multicolumn{2}{c}{\textbf{Deblur}} && \multicolumn{2}{c}{\textbf{SR}}  && \multicolumn{2}{c}{\textbf{Box Inp}} \\
        \cline{3-4} \cline{6-7} \cline{9-10} \cline{12-13}
     Noise & loss & PSNR  & SSIM && PSNR  & SSIM  && PSNR  & SSIM && PSNR  & SSIM \\
   \toprule
\multirow{2}{*}{Poisson} &$\ell_2$ & \textbf{34.82} & \textbf{0.935}  &  & \textbf{26.28} & \textbf{0.709} &  & \textbf{26.26} & \textbf{0.747} &  & \textbf{26.25}  & \textbf{0.889} \\
 & $\ell_1$ & 29.31 & 0.740 &  & 25.70  & 0.679 &  & 24.74 & 0.717 &  & 23.61 & 0.693\\
   \midrule
\multirow{2}{*}{Impulse} & $\ell_2$ & 25.81 & 0.743 &  & 24.12 & 0.627 &  & 20.58 & 0.491 &  & 23.08 & 0.598 \\
  & $\ell_1$ & \textbf{37.94} & \textbf{0.962} &  & \textbf{28.05} & \textbf{0.791} &  & \textbf{24.70}  & \textbf{0.730}  &  & \textbf{25.43} & \textbf{0.908}\\
      \bottomrule
\end{tabular}}%\vspace*{-15pt}
\end{table}

\noindent{\bf Impact of $\ell_1$ vs.\ $\ell_2$ losses for non-Gaussian noise}\ \
To assess the impact of loss function choice on restoration performance under different noise types, we conduct an ablation study using our method with either $\ell_1$ or $\ell_2$ data fidelity terms. We report results on the AFHQ-Cat dataset under Poisson and impulse noise. Table~\ref{tab:afhq} presents the quantitative results. For impulse noise, the $\ell_1$ loss significantly outperforms the $\ell_2$ loss, consistent with the known robustness of $\ell_1$ to sparse noise. Conversely, under Poisson noise, the $\ell_2$ loss yields better performance than $\ell_1$. These results validate our design choice of using loss functions tailored to noise type and highlight the importance of appropriate fidelity terms in inverse problems.

\noindent{\bf Efficiency comparison}\ \
See Table~\ref{tab:time} for the efficiency comparison of these methods. Our method is efficient as the existing methods. Flow-priors is slow due to its sequential approximation. PnP-FBS uses 5 samples to implement the denoising step, while ours use only one sample.
\begin{table}[!htp]
  \centering
  \small
    \fontsize{9}{12}\selectfont 
  \caption{Computation time (in s) to deblur one image. \label{tab:time}} %\vspace*{-10pt}
 \begin{tabular}{@{\hspace{2pt}}c@{\hspace{3pt}}c@{\hspace{3pt}}c@{\hspace{3pt}}c@{\hspace{3pt}}c@{\hspace{3pt}}c@{\hspace{3pt}}c@{\hspace{1pt}}c@{\hspace{1pt}}c@{\hspace{1pt}}c@{\hspace{1pt}}c@{\hspace{1pt}}c@{\hspace{1pt}}c@{\hspace{1pt}}c@{\hspace{1pt}}c@{\hspace{1pt}}c@{\hspace{1pt}}cc@{\hspace{1pt}}c@{\hspace{1pt}}c@{\hspace{0pt}}cc@{\hspace{1pt}}c@{\hspace{1pt}}c@{\hspace{0pt}}cc@{\hspace{1pt}}c@{\hspace{1pt}}c@{\hspace{0pt}}}
 \toprule
   Method & Flow-Priors & OT-ODE & PnP-HQS & PnP-FBS & Ours \\
    \toprule
Time   & 104         & 18     & 12      & 25      & 13 \\
\bottomrule
\end{tabular}\vspace*{-10pt}
\end{table}

\section{Conclusion}

The flow generative model can be integrated into the traditional regularized optimization to address the ill-posedness of imaging inverse problems. Existing plug-and-play (PnP) methods established an iterative proximal splitting based method for smooth squared $\ell_2$ norm data fidelity term induced from Gaussian noise. However, in many practical applications involving non-Gaussian noise, such as Poisson noise and salt-and-pepper impulse noise, the squared $\ell_2$ norm is sub-optimal, and the nonsmooth $\ell_1,\ell_2$-norm based data fidelities are favorable with better performance. To close the gap, we propose integrating the flow generative model into the traditional primal dual hybrid gradient (PDHG) method to exploit the proximity-friendly structure of the $\ell_1,\ell_2$-norm-based losses, and the proximal operator corresponding to the regularization term is implicitly implemented via the denoiser induced by the generative models. 

\section{Acknowledgments}
JL was supported by the National Natural Science Foundation of China (Grant No. 12571472) and supported by the Open Project of Key Laboratory of Mathematics and Information Networks (Beijing University of Posts and Telecommunications), Ministry of Education, China, under Grant No. KF202401).

%\bibliography{main}

\newpage
\clearpage
%\onecolumn
\setcounter{page}{1}
%\newcounter{Theorem}
%\setcounter{Theorem}{1}
%\numberwithin{section}{theorem}
\maketitlesupplementary

\appendix

\section{Conjugate function and proximal operator}\label{sec:ad}

\begin{definition}[\textbf{Conjugate function}]
  Let $f:\mathbb{R}^n\to\mathbb{R}$ be a function. The \emph{conjugate function} of $f$ is the function $f^*:\mathbb{R}^n\to [-\infty,+\infty]$ defined by
  \begin{equation}
    f^*(\bm{y}) = \sup_{\bm{x}\in\mathbb{R}^n}\quad \{\langle\bm{x},\bm{y}\rangle-f(\bm{x})\}.
  \end{equation}
\end{definition}
Note that $f^*$ is convex even if $f$ is not convex.

\begin{definition}[\textbf{Proximal operator}]
  Let $f:\mathbb{R}^n\to\mathbb{R}$ be a function. The \emph{proximal operator} of scaled $\lambda f$ ($\lambda>0$) is the map $\prox_{\lambda f}:\mathbb{R}^n\to\mathbb{R}^n$ defined by
  \begin{equation}
    \prox_{\lambda f}(\bm{y}):=\argmin_{\bm{x}\in\mathbb{R}^n}\quad \{\lambda f(\bm{x}) + \frac{1}{2}\norm{\bm{x}-\bm{y}}_2^2)\}.
  \end{equation}
\end{definition}
Obviously, the proximal operator $\prox_{\lambda f}(\bm{y})$ can be interpreted by projecting of a given point $\bm{y}\in\mathbb{R}^n$ to a point that compromises between minimizing $f$ and being near to $\bm{y}$, and parameter $\lambda$ denotes the trade-off level. Such interpretation leads to an implicit denoising implementation of proximal operator, this is the backstone of the PnP method with a predefined denoiser. %Though the proximal operator needs not be a contraction (unless $f$ is strongly convex), it is \emph{firm nonexpansiveness}, sufficient to guarantee the convergence for the fixed point iteration.

The proximal operator and the conjugate function are connected via the \emph{Moreau decomposition}. More generally, the following relation always holds:
\begin{equation}
  \label{eq:moreau}
  \prox_{\sigma f^*}(\bm{y}) = \bm{y} - \sigma\prox_{\sigma^{-1}f}(\sigma^{-1}\bm{y}).
\end{equation}

\section{More visualizations of restoration on non-Gaussian noise}\label{sec:a1}

\begin{figure*}[!htp]
    \centering %\vspace*{50pt}
    \begin{tabular}{c@{\hspace*{3pt}}c@{\hspace*{1pt}}c@{\hspace*{1pt}}c@{\hspace*{1pt}}c@{\hspace*{1pt}}c@{\hspace*{1pt}}c@{\hspace*{1pt}}c@{\hspace*{1pt}}c@{\hspace*{1pt}}c@{\hspace*{2pt}}c@{\hspace*{2pt}}c@{\hspace*{2pt}}c@{\hspace*{1pt}}}
      \rotatebox[origin=c]{90}{Denoising} &\raisebox{-0.5\height}{\includegraphics[width = 0.13\textwidth]{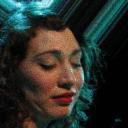}}
& \raisebox{-0.5\height}{\includegraphics[width = 0.13\textwidth]{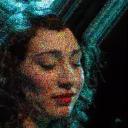}}
& \raisebox{-0.5\height}{\includegraphics[width = 0.13\textwidth]{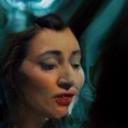}}
& \raisebox{-0.5\height}{\includegraphics[width = 0.13\textwidth]{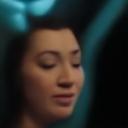}}
& \raisebox{-0.5\height}{\includegraphics[width = 0.13\textwidth]{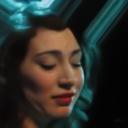}}
& \raisebox{-0.5\height}{\includegraphics[width = 0.13\textwidth]{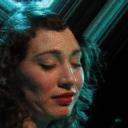}}
& \raisebox{-0.5\height}{\includegraphics[width = 0.13\textwidth]{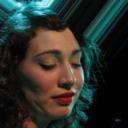}}\\[-2pt]
\rotatebox[origin=c]{90}{Deblur} &\raisebox{-0.5\height}{\includegraphics[width = 0.13\textwidth]{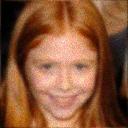}}
& \raisebox{-0.5\height}{\includegraphics[width = 0.13\textwidth]{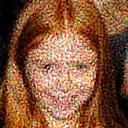}}
& \raisebox{-0.5\height}{\includegraphics[width = 0.13\textwidth]{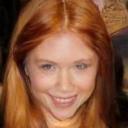}}
& \raisebox{-0.5\height}{\includegraphics[width = 0.13\textwidth]{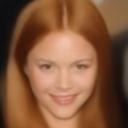}}
& \raisebox{-0.5\height}{\includegraphics[width = 0.13\textwidth]{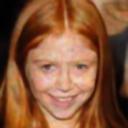}}
& \raisebox{-0.5\height}{\includegraphics[width = 0.13\textwidth]{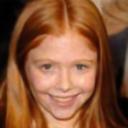}}
& \raisebox{-0.5\height}{\includegraphics[width = 0.13\textwidth]{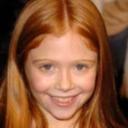}}\\[-2pt]
\rotatebox[origin=c]{90}{SR} &\raisebox{-0.5\height}{\includegraphics[width = 0.13\textwidth]{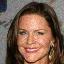}}
& \raisebox{-0.5\height}{\includegraphics[width = 0.13\textwidth]{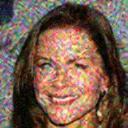}}
& \raisebox{-0.5\height}{\includegraphics[width = 0.13\textwidth]{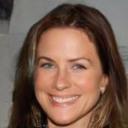}}
& \raisebox{-0.5\height}{\includegraphics[width = 0.13\textwidth]{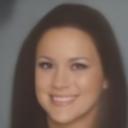}}
& \raisebox{-0.5\height}{\includegraphics[width = 0.13\textwidth]{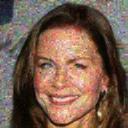}}
& \raisebox{-0.5\height}{\includegraphics[width = 0.13\textwidth]{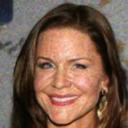}}
& \raisebox{-0.5\height}{\includegraphics[width = 0.13\textwidth]{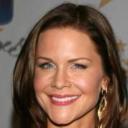}}\\[-2pt]
\rotatebox[origin=c]{90}{Box inp} &\raisebox{-0.5\height}{\includegraphics[width = 0.13\textwidth]{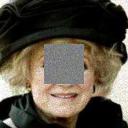}}
& \raisebox{-0.5\height}{\includegraphics[width = 0.13\textwidth]{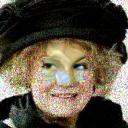}}
& \raisebox{-0.5\height}{\includegraphics[width = 0.13\textwidth]{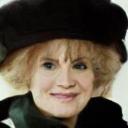}}
& \raisebox{-0.5\height}{\includegraphics[width = 0.13\textwidth]{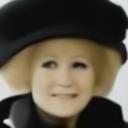}}
& \raisebox{-0.5\height}{\includegraphics[width = 0.13\textwidth]{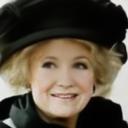}}
& \raisebox{-0.5\height}{\includegraphics[width = 0.13\textwidth]{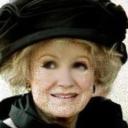}}
      & \raisebox{-0.5\height}{\includegraphics[width = 0.13\textwidth]{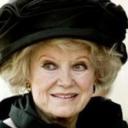}}\\[10pt]
 &  \small{Input} & \small{Flow Priors} & \small{OT-ODE} &\small{PnP-HQS} & \small{PnP-Flow} & \small{Ours} & \small{GT}\\
  \end{tabular}
  \caption{\textbf{Visualization of different methods on four tasks with Poisson noise for CelebA dataset.}\label{fig:poisson_supp} } %\vspace*{-20pt}
  \end{figure*}

\begin{figure*}[!htp]
    \centering %\vspace*{50pt}
    \begin{tabular}{c@{\hspace*{3pt}}c@{\hspace*{1pt}}c@{\hspace*{1pt}}c@{\hspace*{1pt}}c@{\hspace*{1pt}}c@{\hspace*{1pt}}c@{\hspace*{1pt}}c@{\hspace*{1pt}}c@{\hspace*{1pt}}c@{\hspace*{2pt}}c@{\hspace*{2pt}}c@{\hspace*{2pt}}c@{\hspace*{1pt}}}
\rotatebox[origin=c]{90}{Denoising} &\raisebox{-0.5\height}{\includegraphics[width = 0.13\textwidth]{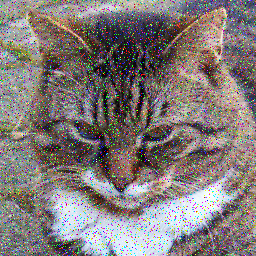}}
& \raisebox{-0.5\height}{\includegraphics[width = 0.13\textwidth]{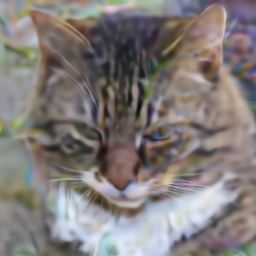}}
& \raisebox{-0.5\height}{\includegraphics[width = 0.13\textwidth]{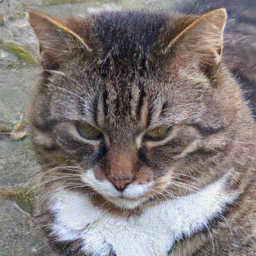}}
& \raisebox{-0.5\height}{\includegraphics[width = 0.13\textwidth]{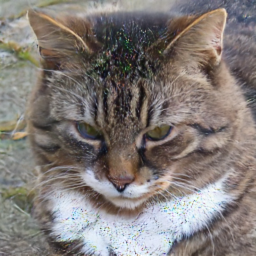}}
& \raisebox{-0.5\height}{\includegraphics[width = 0.13\textwidth]{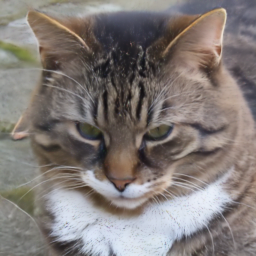}}
& \raisebox{-0.5\height}{\includegraphics[width = 0.13\textwidth]{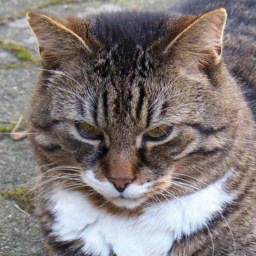}}
  & \raisebox{-0.5\height}{\includegraphics[width = 0.13\textwidth]{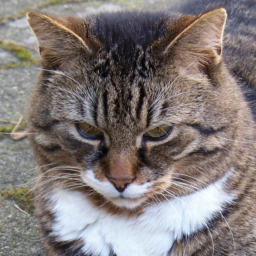}}\\[-2pt]
  \rotatebox[origin=c]{90}{Deblur} &\raisebox{-0.5\height}{\includegraphics[width = 0.13\textwidth]{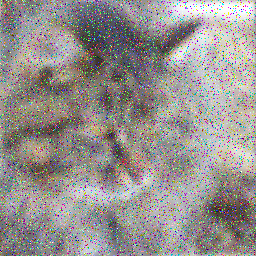}}
& \raisebox{-0.5\height}{\includegraphics[width = 0.13\textwidth]{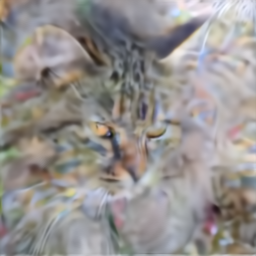}}
& \raisebox{-0.5\height}{\includegraphics[width = 0.13\textwidth]{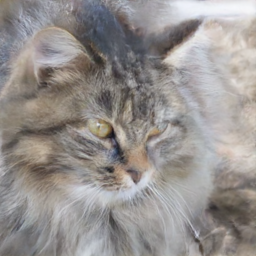}}
& \raisebox{-0.5\height}{\includegraphics[width = 0.13\textwidth]{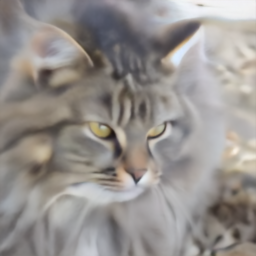}}
& \raisebox{-0.5\height}{\includegraphics[width = 0.13\textwidth]{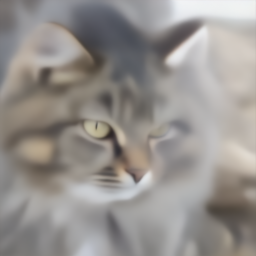}}
& \raisebox{-0.5\height}{\includegraphics[width = 0.13\textwidth]{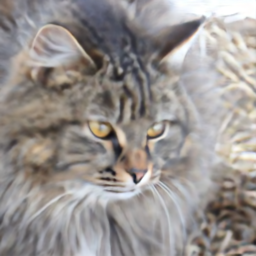}}
& \raisebox{-0.5\height}{\includegraphics[width = 0.13\textwidth]{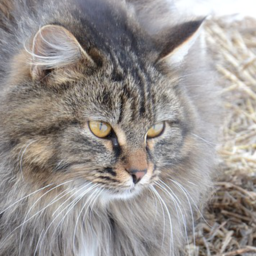}}\\[-2pt]
\rotatebox[origin=c]{90}{SR} &\raisebox{-0.5\height}{\includegraphics[width = 0.13\textwidth]{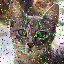}}
& \raisebox{-0.5\height}{\includegraphics[width = 0.13\textwidth]{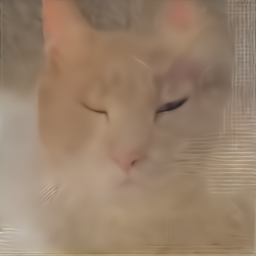}}
& \raisebox{-0.5\height}{\includegraphics[width = 0.13\textwidth]{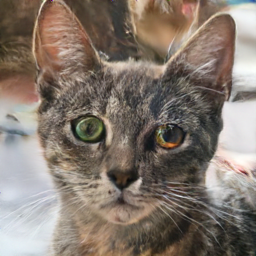}}
& \raisebox{-0.5\height}{\includegraphics[width = 0.13\textwidth]{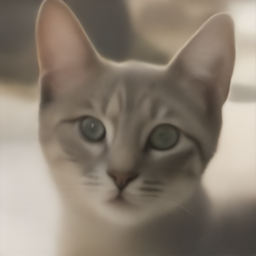}}
& \raisebox{-0.5\height}{\includegraphics[width = 0.13\textwidth]{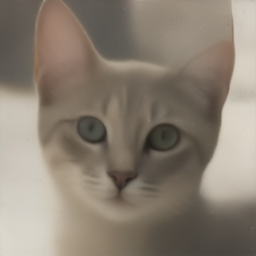}}
& \raisebox{-0.5\height}{\includegraphics[width = 0.13\textwidth]{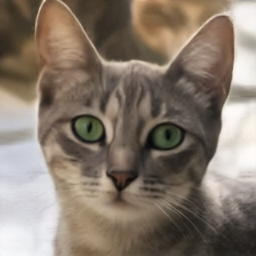}}
  & \raisebox{-0.5\height}{\includegraphics[width = 0.13\textwidth]{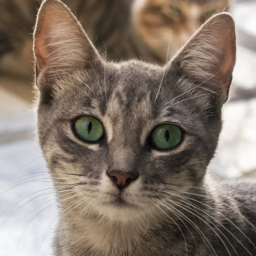}}\\[-2pt]
  \rotatebox[origin=c]{90}{Box inp} &\raisebox{-0.5\height}{\includegraphics[width = 0.13\textwidth]{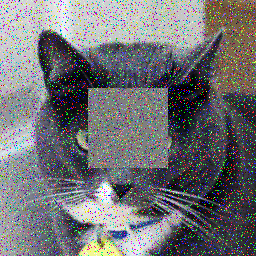}}
& \raisebox{-0.5\height}{\includegraphics[width = 0.13\textwidth]{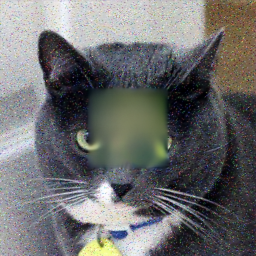}}
& \raisebox{-0.5\height}{\includegraphics[width = 0.13\textwidth]{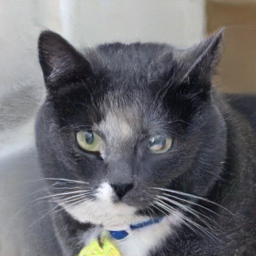}}
& \raisebox{-0.5\height}{\includegraphics[width = 0.13\textwidth]{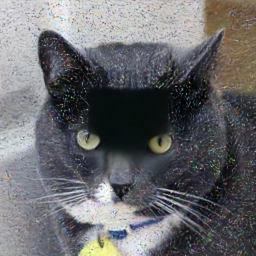}}
& \raisebox{-0.5\height}{\includegraphics[width = 0.13\textwidth]{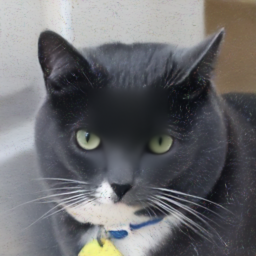}}
& \raisebox{-0.5\height}{\includegraphics[width = 0.13\textwidth]{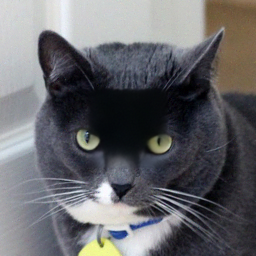}}
& \raisebox{-0.5\height}{\includegraphics[width = 0.13\textwidth]{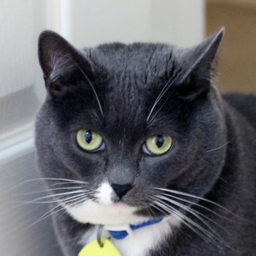}}\\[2pt]
&  \small{Input} & \small{Flow Priors} & \small{OT-ODE} &\small{PnP-HQS} & \small{PnP-Flow} & \small{Ours}  & \small{GT}\\
\end{tabular}
  \caption{\textbf{Visualization of different methods on four tasks with impulse noise for AFHQ-Cat.}\label{fig:sp_supp}}
\end{figure*}

We present more visualizations of restoration for the four considered imaging tasks on Poisson and salt-and-pepper noise. See Figures~\ref{fig:poisson_supp} and~\ref{fig:sp_supp} for the visualization of different methods on the four tasks with Poisson noise for CelebA dataset and salt-and-pepper noise for AFHQ-Cat dataset. Compared to other methods, ours reconstructs more faithful outputs with better details. The visualization matches the quantitative results in the tables of the main paper.

\section{Results for diffusion model}\label{sec:ae}

\begin{figure*}[!htp]
    \centering %\vspace*{50pt}
    \begin{tabular}{c@{\hspace*{3pt}}c@{\hspace*{1pt}}c@{\hspace*{1pt}}c@{\hspace*{1pt}}c@{\hspace*{1pt}}c@{\hspace*{1pt}}c@{\hspace*{1pt}}c@{\hspace*{1pt}}c@{\hspace*{1pt}}c@{\hspace*{2pt}}c@{\hspace*{2pt}}c@{\hspace*{2pt}}c@{\hspace*{1pt}}}
\rotatebox[origin=c]{90}{Denoising} &\raisebox{-0.5\height}{\includegraphics[width = 0.14\textwidth]{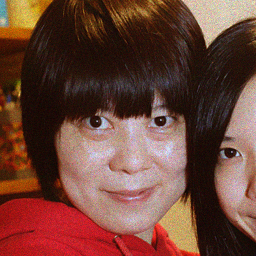}}
& \raisebox{-0.5\height}{\includegraphics[width = 0.14\textwidth]{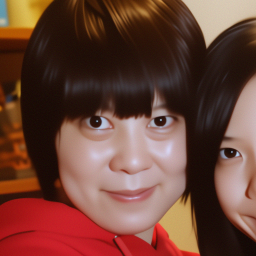}}
& \raisebox{-0.5\height}{\includegraphics[width = 0.14\textwidth]{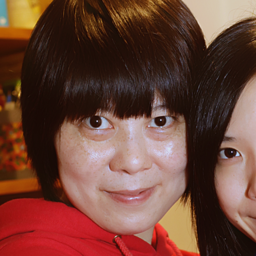}}
&\raisebox{-0.5\height}{\includegraphics[width = 0.14\textwidth]{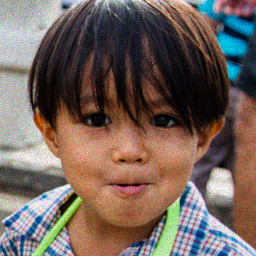}}
& \raisebox{-0.5\height}{\includegraphics[width = 0.14\textwidth]{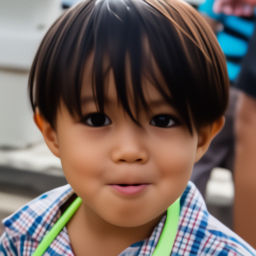}}
& \raisebox{-0.5\height}{\includegraphics[width = 0.14\textwidth]{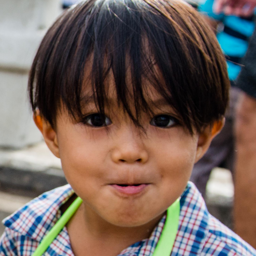}}\\[-2pt]
\rotatebox[origin=c]{90}{Denoising} &\raisebox{-0.5\height}{\includegraphics[width = 0.14\textwidth]{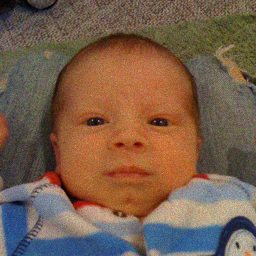}}
& \raisebox{-0.5\height}{\includegraphics[width = 0.14\textwidth]{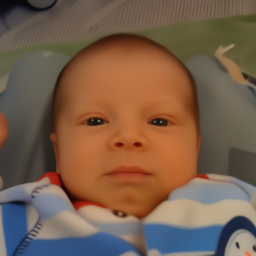}}
& \raisebox{-0.5\height}{\includegraphics[width = 0.14\textwidth]{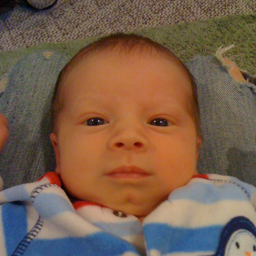}}&
\raisebox{-0.5\height}{\includegraphics[width = 0.14\textwidth]{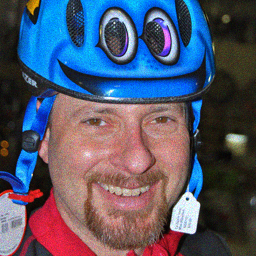}}
& \raisebox{-0.5\height}{\includegraphics[width = 0.14\textwidth]{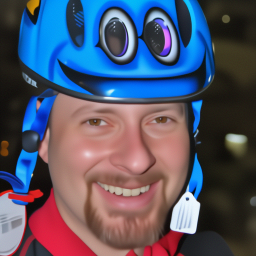}}
& \raisebox{-0.5\height}{\includegraphics[width = 0.14\textwidth]{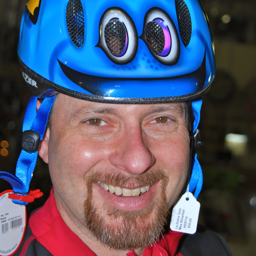}}\\[-2pt]
\rotatebox[origin=c]{90}{Deblur} &\raisebox{-0.5\height}{\includegraphics[width = 0.14\textwidth]{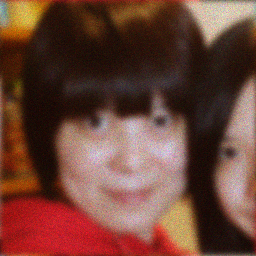}}
& \raisebox{-0.5\height}{\includegraphics[width = 0.14\textwidth]{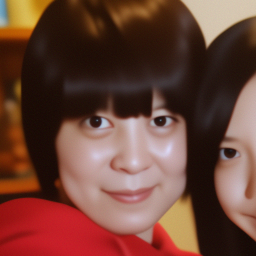}}
& \raisebox{-0.5\height}{\includegraphics[width = 0.14\textwidth]{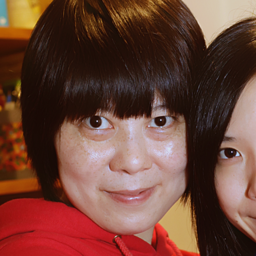}}
&\raisebox{-0.5\height}{\includegraphics[width = 0.14\textwidth]{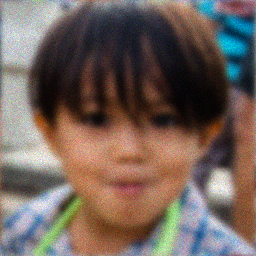}}
& \raisebox{-0.5\height}{\includegraphics[width = 0.14\textwidth]{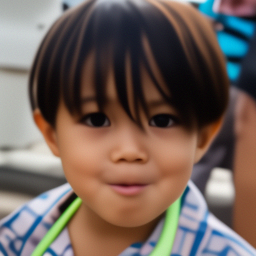}}
& \raisebox{-0.5\height}{\includegraphics[width = 0.14\textwidth]{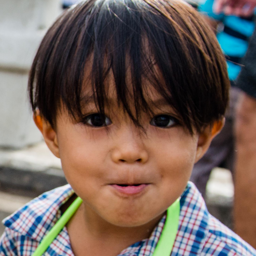}}\\[-2pt]
\rotatebox[origin=c]{90}{Deblur} &\raisebox{-0.5\height}{\includegraphics[width = 0.14\textwidth]{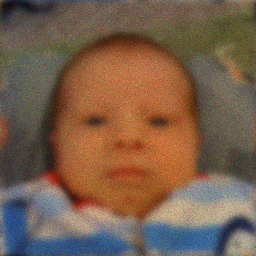}}
& \raisebox{-0.5\height}{\includegraphics[width = 0.14\textwidth]{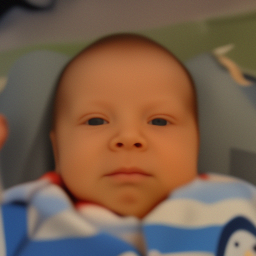}}
& \raisebox{-0.5\height}{\includegraphics[width = 0.14\textwidth]{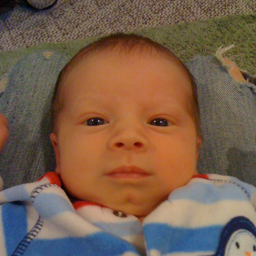}}&
\raisebox{-0.5\height}{\includegraphics[width = 0.14\textwidth]{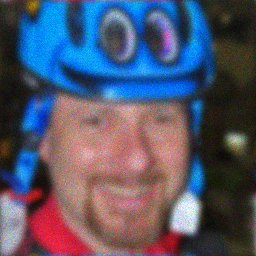}}
& \raisebox{-0.5\height}{\includegraphics[width = 0.14\textwidth]{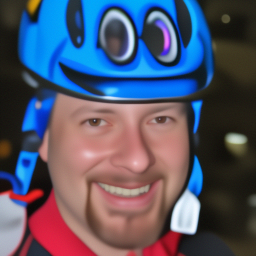}}
& \raisebox{-0.5\height}{\includegraphics[width = 0.14\textwidth]{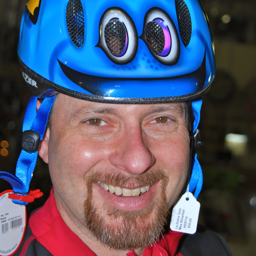}}\\[-2pt]
\rotatebox[origin=c]{90}{SR} &\raisebox{-0.5\height}{\includegraphics[width = 0.14\textwidth]{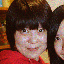}}
& \raisebox{-0.5\height}{\includegraphics[width = 0.14\textwidth]{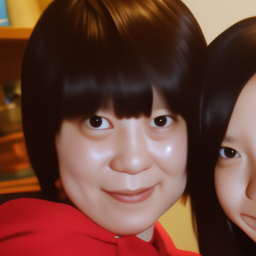}}
& \raisebox{-0.5\height}{\includegraphics[width = 0.14\textwidth]{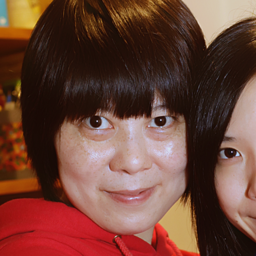}}
&\raisebox{-0.5\height}{\includegraphics[width = 0.14\textwidth]{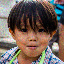}}
& \raisebox{-0.5\height}{\includegraphics[width = 0.14\textwidth]{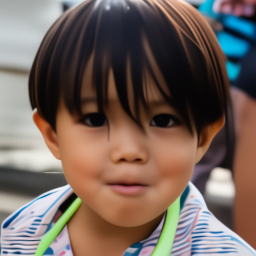}}
& \raisebox{-0.5\height}{\includegraphics[width = 0.14\textwidth]{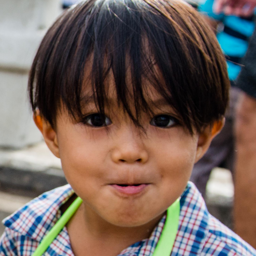}}\\[-2pt]
\rotatebox[origin=c]{90}{SR} &\raisebox{-0.5\height}{\includegraphics[width = 0.14\textwidth]{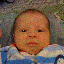}}
& \raisebox{-0.5\height}{\includegraphics[width = 0.14\textwidth]{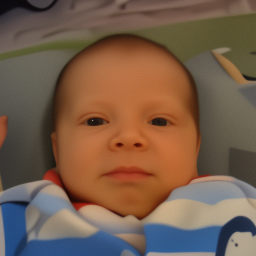}}
& \raisebox{-0.5\height}{\includegraphics[width = 0.14\textwidth]{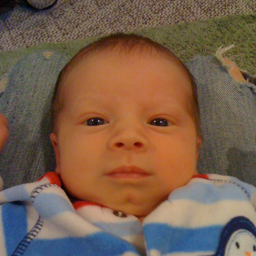}}&
\raisebox{-0.5\height}{\includegraphics[width = 0.14\textwidth]{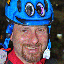}}
& \raisebox{-0.5\height}{\includegraphics[width = 0.14\textwidth]{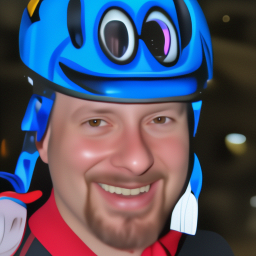}}
& \raisebox{-0.5\height}{\includegraphics[width = 0.14\textwidth]{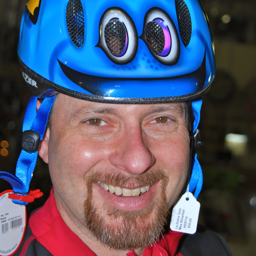}}\\[-2pt]
\rotatebox[origin=c]{90}{Box inp} &\raisebox{-0.5\height}{\includegraphics[width = 0.14\textwidth]{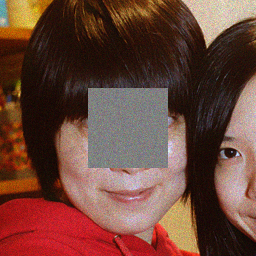}}
& \raisebox{-0.5\height}{\includegraphics[width = 0.14\textwidth]{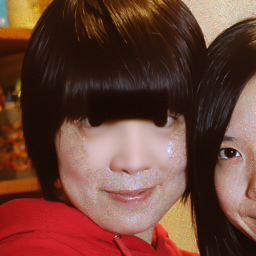}}
& \raisebox{-0.5\height}{\includegraphics[width = 0.14\textwidth]{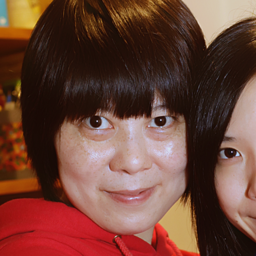}}
&\raisebox{-0.5\height}{\includegraphics[width = 0.14\textwidth]{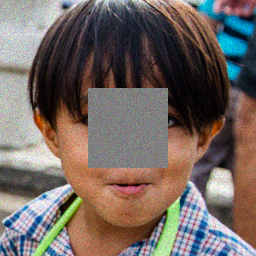}}
& \raisebox{-0.5\height}{\includegraphics[width = 0.14\textwidth]{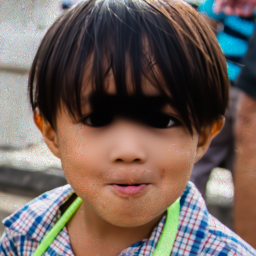}}
& \raisebox{-0.5\height}{\includegraphics[width = 0.14\textwidth]{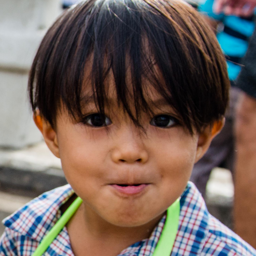}}\\[-2pt]
\rotatebox[origin=c]{90}{Box inp} &\raisebox{-0.5\height}{\includegraphics[width = 0.14\textwidth]{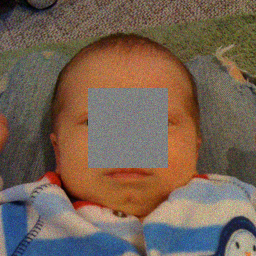}}
& \raisebox{-0.5\height}{\includegraphics[width = 0.14\textwidth]{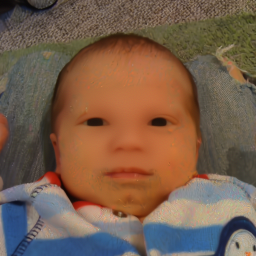}}
& \raisebox{-0.5\height}{\includegraphics[width = 0.14\textwidth]{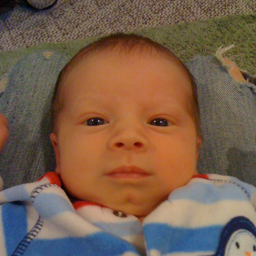}}&
\raisebox{-0.5\height}{\includegraphics[width = 0.14\textwidth]{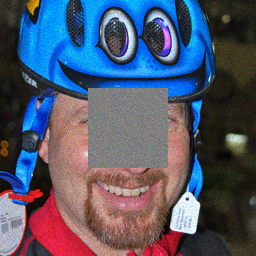}}
& \raisebox{-0.5\height}{\includegraphics[width = 0.14\textwidth]{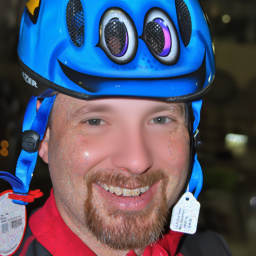}}
& \raisebox{-0.5\height}{\includegraphics[width = 0.14\textwidth]{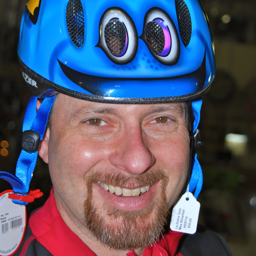}}\\[-2pt]
 &  \small{Input} & \small{Ours} & \small{GT} &  \small{Input} & \small{Ours} & \small{GT} \\
  \end{tabular}
  \caption{\textbf{Visualization of our method on four tasks with Poisson noise for FFHQ dataset.}\label{fig:ffhq_supp} } %\vspace*{-20pt}
  \end{figure*}

\begin{figure*}[!htp]
  \centering %\vspace*{50pt}
  \begin{tabular}{c@{\hspace*{3pt}}c@{\hspace*{1pt}}c@{\hspace*{1pt}}c@{\hspace*{1pt}}c@{\hspace*{1pt}}c@{\hspace*{1pt}}c@{\hspace*{1pt}}c@{\hspace*{1pt}}c@{\hspace*{1pt}}c@{\hspace*{2pt}}c@{\hspace*{2pt}}c@{\hspace*{2pt}}c@{\hspace*{1pt}}}
\rotatebox[origin=c]{90}{Denoising} &\raisebox{-0.5\height}{\includegraphics[width = 0.14\textwidth]{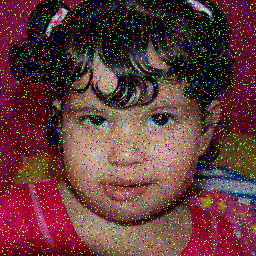}}
& \raisebox{-0.5\height}{\includegraphics[width = 0.14\textwidth]{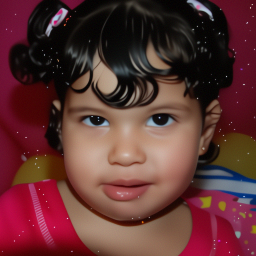}}
& \raisebox{-0.5\height}{\includegraphics[width = 0.14\textwidth]{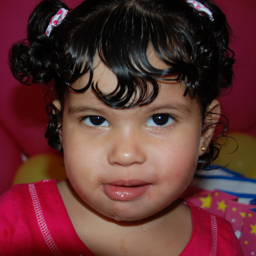}}
&\raisebox{-0.5\height}{\includegraphics[width = 0.14\textwidth]{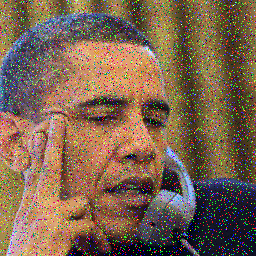}}
& \raisebox{-0.5\height}{\includegraphics[width = 0.14\textwidth]{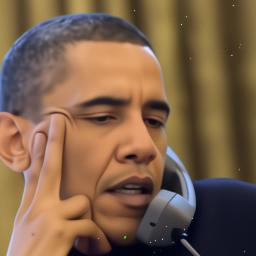}}
& \raisebox{-0.5\height}{\includegraphics[width = 0.14\textwidth]{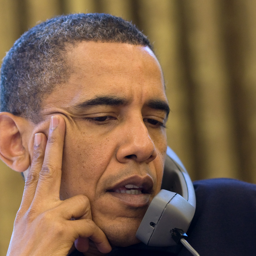}}\\[-2pt]
\rotatebox[origin=c]{90}{Denoising} &\raisebox{-0.5\height}{\includegraphics[width = 0.14\textwidth]{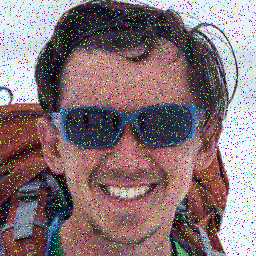}}
& \raisebox{-0.5\height}{\includegraphics[width = 0.14\textwidth]{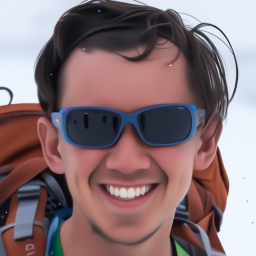}}
& \raisebox{-0.5\height}{\includegraphics[width = 0.14\textwidth]{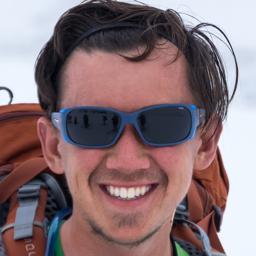}}&
\raisebox{-0.5\height}{\includegraphics[width = 0.14\textwidth]{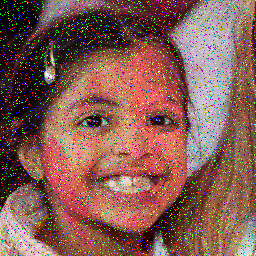}}
& \raisebox{-0.5\height}{\includegraphics[width = 0.14\textwidth]{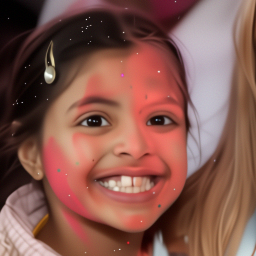}}
& \raisebox{-0.5\height}{\includegraphics[width = 0.14\textwidth]{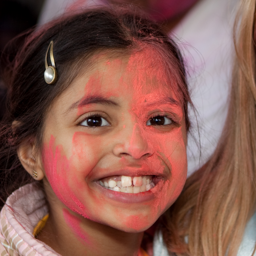}}\\[-2pt]
\rotatebox[origin=c]{90}{Deblur} &\raisebox{-0.5\height}{\includegraphics[width = 0.14\textwidth]{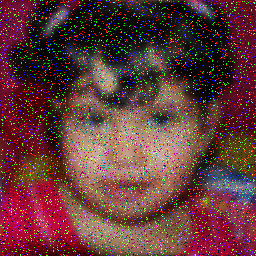}}
& \raisebox{-0.5\height}{\includegraphics[width = 0.14\textwidth]{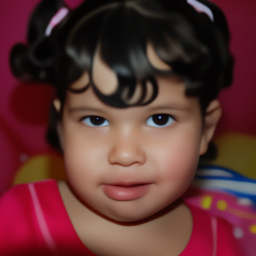}}
& \raisebox{-0.5\height}{\includegraphics[width = 0.14\textwidth]{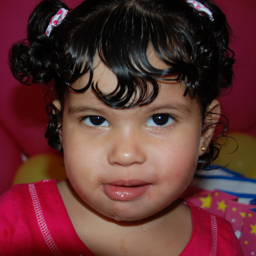}}
&\raisebox{-0.5\height}{\includegraphics[width = 0.14\textwidth]{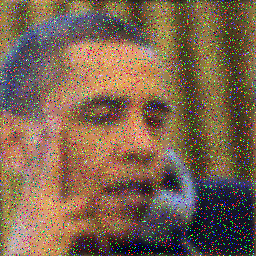}}
& \raisebox{-0.5\height}{\includegraphics[width = 0.14\textwidth]{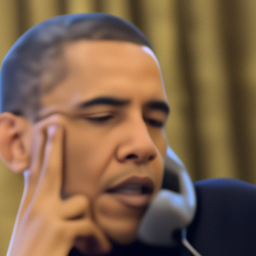}}
& \raisebox{-0.5\height}{\includegraphics[width = 0.14\textwidth]{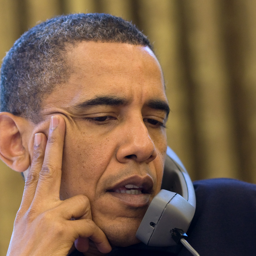}}\\[-2pt]
\rotatebox[origin=c]{90}{Deblur} &\raisebox{-0.5\height}{\includegraphics[width = 0.14\textwidth]{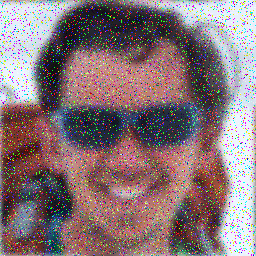}}
& \raisebox{-0.5\height}{\includegraphics[width = 0.14\textwidth]{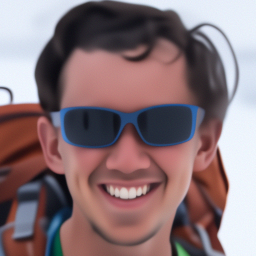}}
& \raisebox{-0.5\height}{\includegraphics[width = 0.14\textwidth]{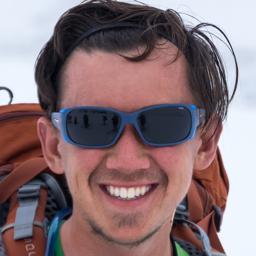}}&
\raisebox{-0.5\height}{\includegraphics[width = 0.14\textwidth]{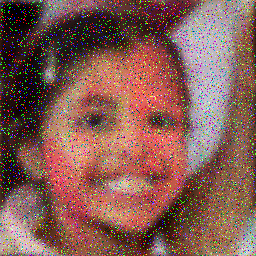}}
& \raisebox{-0.5\height}{\includegraphics[width = 0.14\textwidth]{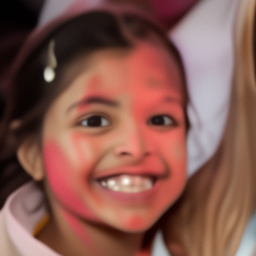}}
& \raisebox{-0.5\height}{\includegraphics[width = 0.14\textwidth]{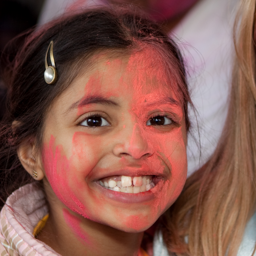}}\\[-2pt]
\rotatebox[origin=c]{90}{SR} &\raisebox{-0.5\height}{\includegraphics[width = 0.14\textwidth]{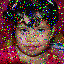}}
& \raisebox{-0.5\height}{\includegraphics[width = 0.14\textwidth]{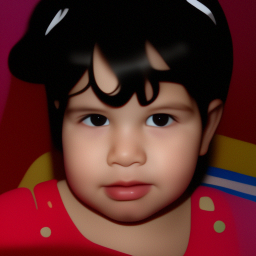}}
& \raisebox{-0.5\height}{\includegraphics[width = 0.14\textwidth]{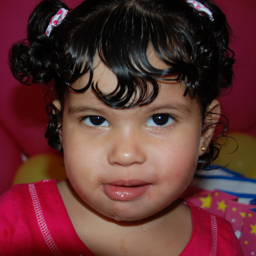}}
&\raisebox{-0.5\height}{\includegraphics[width = 0.14\textwidth]{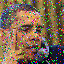}}
& \raisebox{-0.5\height}{\includegraphics[width = 0.14\textwidth]{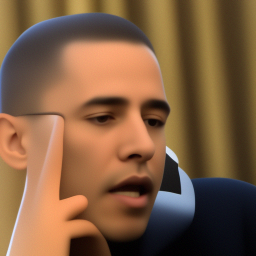}}
& \raisebox{-0.5\height}{\includegraphics[width = 0.14\textwidth]{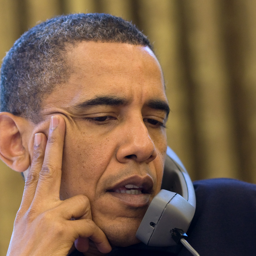}}\\[-2pt]
\rotatebox[origin=c]{90}{SR} &\raisebox{-0.5\height}{\includegraphics[width = 0.14\textwidth]{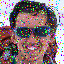}}
& \raisebox{-0.5\height}{\includegraphics[width = 0.14\textwidth]{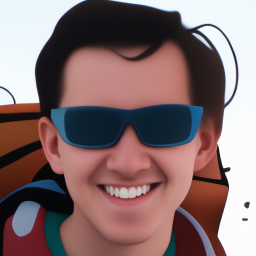}}
& \raisebox{-0.5\height}{\includegraphics[width = 0.14\textwidth]{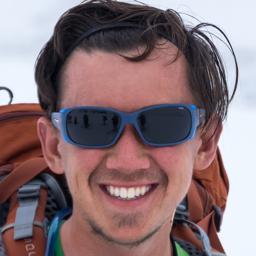}}&
\raisebox{-0.5\height}{\includegraphics[width = 0.14\textwidth]{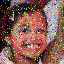}}
& \raisebox{-0.5\height}{\includegraphics[width = 0.14\textwidth]{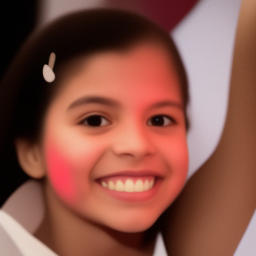}}
& \raisebox{-0.5\height}{\includegraphics[width = 0.14\textwidth]{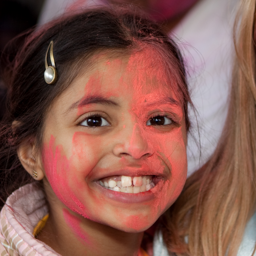}}\\[-2pt]
\rotatebox[origin=c]{90}{Box inp} &\raisebox{-0.5\height}{\includegraphics[width = 0.14\textwidth]{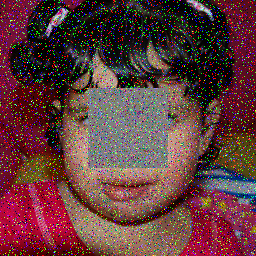}}
& \raisebox{-0.5\height}{\includegraphics[width = 0.14\textwidth]{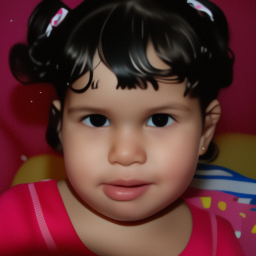}}
& \raisebox{-0.5\height}{\includegraphics[width = 0.14\textwidth]{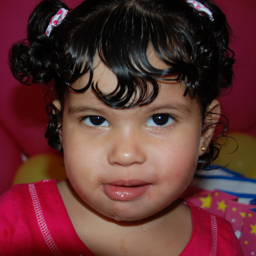}}
&\raisebox{-0.5\height}{\includegraphics[width = 0.14\textwidth]{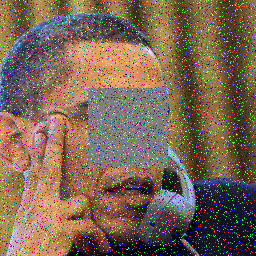}}
& \raisebox{-0.5\height}{\includegraphics[width = 0.14\textwidth]{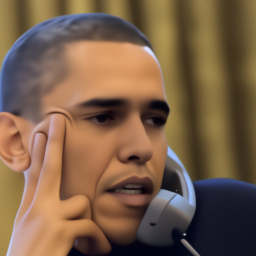}}
& \raisebox{-0.5\height}{\includegraphics[width = 0.14\textwidth]{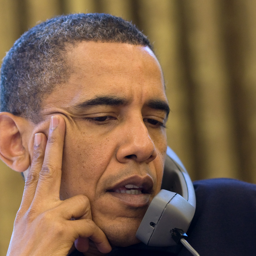}}\\[-2pt]
\rotatebox[origin=c]{90}{Box inp} &\raisebox{-0.5\height}{\includegraphics[width = 0.14\textwidth]{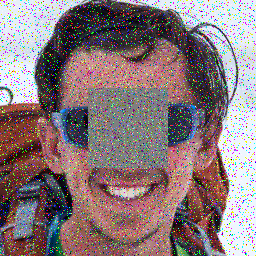}}
& \raisebox{-0.5\height}{\includegraphics[width = 0.14\textwidth]{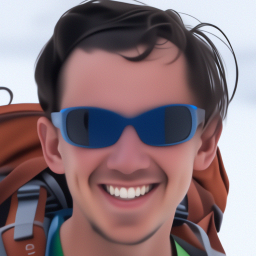}}
& \raisebox{-0.5\height}{\includegraphics[width = 0.14\textwidth]{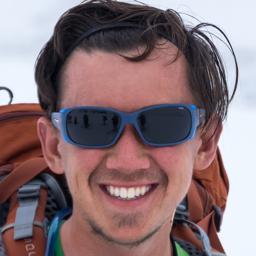}}&
\raisebox{-0.5\height}{\includegraphics[width = 0.14\textwidth]{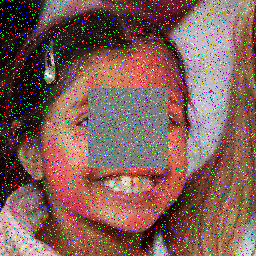}}
& \raisebox{-0.5\height}{\includegraphics[width = 0.14\textwidth]{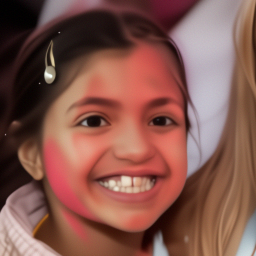}}
& \raisebox{-0.5\height}{\includegraphics[width = 0.14\textwidth]{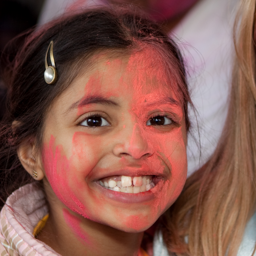}}\\[-2pt]
&  \small{Input} & \small{Ours} & \small{GT} &  \small{Input} & \small{Ours} & \small{GT} \\
\end{tabular}
\caption{\textbf{Visualization of our method on four tasks with impulse noise for FFHQ dataset.}\label{fig:ffhq_sp_supp} } %\vspace*{-20pt}
\end{figure*}

For the diffusion model, we consider the FFHQ dataset, with image size $256\times 256$ for its available open source pretrained model \verb+‘ffhq_10m.pt’+ from DPS implementation code. 100 test images are randomly selected. We evaluate the performance of our method on the considered four tasks with Poisson noise and salt-and-pepper noise. The considered noise level is the same as the main paper. We provide the quantitative results for the two noise types and the visualization of the restorations. See Table~\ref{tab:ffhq} for the quantitative results of our method on the four tasks for the two considered non-Gaussian noise types. See Figures~\ref{fig:ffhq_supp} and~\ref{fig:ffhq_sp_supp} for the visualization of our restored images. It shows that our method works well for non-Gaussian noise.

\begin{table}[!htp]
  \centering
  \small
    %\vspace*{-10pt}
    \fontsize{9}{12}\selectfont 
  \caption{Quantitative results of our method for the four tasks on the FFHQ dataset with non-Gaussian noise. \label{tab:ffhq}}
 \begin{tabular}{@{\hspace{0pt}}c@{\hspace{1pt}}c@{\hspace{1pt}}c@{\hspace{1pt}}c@{\hspace{1pt}}c@{\hspace{1pt}}c@{\hspace{1pt}}c@{\hspace{1pt}}c@{\hspace{1pt}}c@{\hspace{1pt}}c@{\hspace{1pt}}c@{\hspace{1pt}}c@{\hspace{1pt}}c@{\hspace{1pt}}c@{\hspace{1pt}}c@{\hspace{1pt}}c@{\hspace{1pt}}cc@{\hspace{1pt}}c@{\hspace{1pt}}c@{\hspace{0pt}}cc@{\hspace{1pt}}c@{\hspace{1pt}}c@{\hspace{0pt}}cc@{\hspace{1pt}}c@{\hspace{1pt}}c@{\hspace{0pt}}}
 \toprule
  FFHQ  &   \multicolumn{2}{c}{\textbf{Denoising}}  && \multicolumn{2}{c}{\textbf{Deblur}} && \multicolumn{2}{c}{\textbf{SR}}  && \multicolumn{2}{c}{\textbf{Box Inp}} \\
        \cline{2-3} \cline{5-6} \cline{8-9} \cline{11-12}
     Noise & PSNR  & SSIM && PSNR  & SSIM  && PSNR  & SSIM && PSNR  & SSIM \\
   \toprule
Poisson & 32.16 & 0.874 &  & 27.92 & 0.783 &  & 26.64 & 0.790  &  & 28.77 & 0.868 \\
SP      & 31.56 & 0.867 &  & 28.51 & 0.801 &  & 20.12 & 0.598 &  & 21.82 & 0.569\\
   \bottomrule
\end{tabular}
\end{table}

\section{Results for rectified flow  model}\label{sec:af}

For rectified flow model, the CelebAHQ dataset with image size $256\times 256$ is considered, 100 test images are selected from the original validation dataset. The pretrained model is from the original rectified flow implementation code. For the rectified flow model, the stepsize $\tau$ is time-dependent with the form $(1-t_k)^\alpha$ with $\alpha$ tuned for different tasks and datasets. For all the considered four tasks, we set $\tau=0.8$.

We still evaluate our method and the compared methods for the non-Gaussian noise. We present the quantitative results in Table~\ref{tab:poisson_hq}. It shows the promising performance of our method over the existing methods designed for Gaussian noise. 

\begin{table}[!htp]
  \centering
  \small
    %\vspace*{-10pt}
    \fontsize{9}{12}\selectfont 
  \caption{Quantitative results of different methods for different tasks on the CelebAHQ datasets with non-Gaussian noise. \label{tab:poisson_hq}}
 \begin{tabular}{@{\hspace{0pt}}c@{\hspace{1pt}}c@{\hspace{1pt}}c@{\hspace{1pt}}c@{\hspace{1pt}}c@{\hspace{1pt}}c@{\hspace{1pt}}c@{\hspace{1pt}}c@{\hspace{1pt}}c@{\hspace{1pt}}c@{\hspace{1pt}}c@{\hspace{1pt}}c@{\hspace{1pt}}c@{\hspace{1pt}}c@{\hspace{1pt}}c@{\hspace{1pt}}c@{\hspace{1pt}}cc@{\hspace{1pt}}c@{\hspace{1pt}}c@{\hspace{0pt}}cc@{\hspace{1pt}}c@{\hspace{1pt}}c@{\hspace{0pt}}cc@{\hspace{1pt}}c@{\hspace{1pt}}c@{\hspace{0pt}}}
 \toprule
  Poisson  &   \multicolumn{2}{c}{\textbf{Denoising}}  && \multicolumn{2}{c}{\textbf{Deblur}} && \multicolumn{2}{c}{\textbf{SR}}  && \multicolumn{2}{c}{\textbf{Box Inp}} \\
        \cline{2-3} \cline{5-6} \cline{8-9} \cline{11-12}
     Method & PSNR  & SSIM && PSNR  & SSIM  && PSNR  & SSIM && PSNR  & SSIM \\
   \toprule
  Flow-Priors & 28.59 & 0.810  &  & 27.80  & 0.773 &  & 13.50  & 0.385 &  & 24.04 & 0.768 \\
OT-ODE      & 24.85 & 0.566 &  & 22.45 & 0.478 &  & 18.41 & 0.391 &  & 22.24 & 0.535 \\
PnP-HQS     & 21.54 & 0.658 &  & 20.72 & 0.632 &  & 21.96 & 0.675 &  & 20.63 & 0.642 \\
PnP-FBS     & 31.48 & 0.901 &  & 27.47 & 0.814 &  & 24.19 & 0.649 &  & 22.72 & 0.553 \\
Ours        & \textbf{37.41} & \textbf{0.948} &  & \textbf{29.86} & \textbf{0.851} &  & \textbf{28.68} & \textbf{0.844} &  & \textbf{27.67} & \textbf{0.912}\\
   \bottomrule
\toprule
 Impulse   &   \multicolumn{2}{c}{\textbf{Denoising}}  && \multicolumn{2}{c}{\textbf{Deblur}} && \multicolumn{2}{c}{\textbf{SR}}  && \multicolumn{2}{c}{\textbf{Box Inp}} \\
        \cline{2-3} \cline{5-6} \cline{8-9} \cline{11-12}
     Method & PSNR  & SSIM && PSNR  & SSIM  && PSNR  & SSIM && PSNR  & SSIM \\
   \toprule
  Flow-Priors & 29.43 & 0.806 &  & 21.35 & 0.720  &  & 13.26 & 0.379 &  & 17.75 & 0.752 \\
OT-ODE      & 23.30  & 0.443 &  & 22.26 & 0.452 &  & 19.11 & 0.290  &  & 21.54 & 0.419 \\
PnP-HQS     & 26.42 & 0.788 &  & 24.76 & 0.728 &  & 21.98 & 0.579 &  & 24.27 & 0.760  \\
PnP-FBS     & 24.42 & 0.599 &  & 25.60  & 0.754 &  & 21.13 & 0.571 &  & 23.86 & 0.751 \\
Ours        & \textbf{36.35} & \textbf{0.917} &  & \textbf{32.89} & \textbf{0.907} &  & \textbf{26.37} & \textbf{0.846} &  & \textbf{27.52} & \textbf{0.887} \\
\bottomrule
\end{tabular}
\end{table}

\newpage

  %\vspace*{-18pt}

\end{document}